\documentclass[conference]{IEEEtran}
\IEEEoverridecommandlockouts

\usepackage{booktabs}

\usepackage{amsmath}

\usepackage{algorithm}
\usepackage{algpseudocode}
\usepackage{todonotes}

\usepackage{multirow}
\usepackage{longtable}

\usepackage{subcaption}
\usepackage{alphalph}
\usepackage{graphicx}

\usepackage{xr}
\usepackage{textcomp}
\usepackage{xcolor}
\usepackage[justification=centering]{caption}
\usepackage{url}
\usepackage{hyperref}

\usepackage{makecell}
\usepackage{booktabs}
\usepackage{hyperref}
\usepackage{url}

\usepackage{comment}
\usepackage{adjustbox}
\usepackage{textcomp}
\usepackage{xcolor}

\usepackage{enumerate}
\usepackage[skip=4pt]{caption}

\usepackage{lastpage}
\usepackage{fancyhdr}
\usepackage[switch]{lineno}

\def\BibTeX{{\rm B\kern-.05em{\sc i\kern-.025em b}\kern-.08em
    T\kern-.1667em\lower.7ex\hbox{E}\kern-.125emX}}

\DeclareRobustCommand*\circled[1]{\tikz[teal,baseline=(char.base)]{
            \node[shape=circle,fill,inner sep=1.3pt] (char) {\textcolor{white}{#1}};}}

\newtheorem{theorem}{Theorem}[section]

\begin{document}
\title{Adaptive Patching for High-resolution Image Segmentation with Transformers}


\author{\IEEEauthorblockN{Enzhi Zhang}
\IEEEauthorblockA{\textit{Grad. School of Info. Science and Technology
} \\
\textit{Hokkaido University}\\
Sapporo, Japan\\
enzhi.zhang.n6@elms.hokudai.ac.jp
}
\and
\IEEEauthorblockN{Isaac Lyngaas}
\IEEEauthorblockA{\textit{Oak Ridge National Laboratory
} \\
Tennessee, United States of America\\
lyngaasir@ornl.gov
}
\and
\IEEEauthorblockN{Peng Chen}
\IEEEauthorblockA{\textit{National Institute of Advanced Industrial Science and Technology AIST
} \\
Tokyo, Japan\\
chin.hou@aist.go.jp
}
\and
\IEEEauthorblockN{Xiao Wang}
\IEEEauthorblockA{\textit{Oak Ridge National Laboratory
} \\
Tennessee, United States of America\\
wangx2@ornl.gov
}
\and
\IEEEauthorblockN{Jun Igarashi}
\IEEEauthorblockA{\textit{RIKEN Center for Computational Science
} \\
Tokyo, Japan\\
jigarashi@riken.jp
}
\and
\IEEEauthorblockN{Yuankai Huo}
\IEEEauthorblockA{\textit{Vanderbilt University
} \\
Tennessee, United States of America\\
yuankai.huo@vanderbilt.edu
}
\and
\IEEEauthorblockN{Mohamed Wahib}
\IEEEauthorblockA{\textit{RIKEN Center for Computational Science} \\
Kobe, Japan \\
mohamed.attia@riken.jp}
\and
\IEEEauthorblockN{Masaharu Munetomo}
\IEEEauthorblockA{\textit{Information Initiative Center} \\
\textit{Hokkaido University}\\
Sapporo, Japan \\
munetomo@iic.hokudai.ac.jp}
}
    
\maketitle

\begin{abstract}
Attention-based models are proliferating in the space of image analytics, including segmentation. The standard method of feeding images to transformer encoders is to divide the images into patches and then feed the patches to the model as a linear sequence of tokens. For high-resolution images, e.g. microscopic pathology images, the quadratic compute and memory cost prohibits the use of an attention-based model, if we are to use smaller patch sizes that are favorable in segmentation. The solution is to either use custom complex multi-resolution models or approximate attention schemes. We take inspiration from Adapative Mesh Refinement (AMR) methods in HPC by adaptively patching the images, as a pre-processing step, based on the image details to reduce the number of patches being fed to the model, by orders of magnitude. This method has a negligible overhead, and works seamlessly with any attention-based model, i.e. it is a pre-processing step that can be adopted by any attention-based model without friction. We demonstrate superior segmentation quality over SoTA segmentation models for real-world pathology datasets while gaining a geomean speedup of $6.9\times$ for resolutions up to $64K^2$, on up to $2,048$ GPUs.
\end{abstract}


\section{Introduction}
Recently, Vision Transformers (ViTs) have emerged as a transformative paradigm in computer vision, demonstrating remarkable success in image classification tasks~\cite{vaswani2017attention, dosovitskiy2021image, hu2021pyramid, wang2021deep, carion2020end}. To effectively tackle dense prediction tasks like segmentation, numerous efforts have introduced variations on ViTs~\cite{thisanke2023semantic,liu2021swin,li2021vit,chen2021crossvit}. Others have explored combinations of transformers with U-Net like architectures~\cite{liu2021swin, hatamizadeh2022unetr, chen2021transunet, shamshad2023transformers}. However, employing ViTs with high-resolution images presents distinct scalability challenges, especially when processing small-sized image patches arranged into long sequences of complex visual data in medical imaging~\cite{hatamizadeh2022unetr, strudel2021segmenter}.


The challenge of handling long sequences in ViTs arises from the quadratic computational complexity associated with self-attention mechanism, leading to significant computational demands~\cite{parmar2018image}. Consequently, traditional ViTs encounter limitations when applied to high-resolution images, where detailed information spans extensive spatial contexts.

There are other two approaches that address the long sequence scaling problem, yet they do not reduce the total amount of work. The Sequence parallelism approach distributes long sequences into sequence segments among workers (GPUs): Deep-Speed Ulysses~\cite{jacobs2023deepspeed}, LightSeq~\cite{li2023lightseq}, RingAttention~\cite{liu2023ring}, and LLS~\cite{wang2023ultralong}. The blocking/titling approach aims to tile the attention matrix into sub-matrices that fit into the user-managed cache memory: FlashAttention 1~\cite{FlashAttention22} and 2~\cite{dao2023flashattention2}, and the Swin transformer~\cite{liu2021swin} that adopts a shifted windowing technique, breaking down the image into smaller overlapping windows for processing within the transformer. The blocking/tiling approach allows scaling for longer sequences to the available memory, however, the total amount of compute is not reduced.

In contrast, there are two other approaches that address the long sequence scaling problem by reducing the amount of work (not necessarily specific to vision transformers): attention approximation and hierarchical training. 

The approximation attention approaches approximate the self-attention mechanism through spectral attention~\cite{pmlr-v162-dao22a,bo2023specformer,NEURIPS2021_b4fd1d2c}, low-rank approximation~\cite{Choromanski20,Katharopoulos20}, sparse attention matrix sampling~\cite{sparse-transformer19,reformer20,Roy21,beltagy2020longformer,bigbird20}, infrequent self-attention updates~\cite{ying2021lazyformer,rabe2022selfattention}, or their combinations~\cite{chen2021scatterbrain}. 
Approximation methods greatly reduce the memory and computation cost, with some reducing the quadratic complexity of self-attention to be linear. Yet, loss of information due to approximating the self-attention could have negative impact on accuracy, especially for long-range sequences. Experiments show a notable drop in accuracy when the compression ratio surpasses 70\%~\cite{Shi21}. Finally, implementing approximation approaches is complex and often requires custom operators and sparse formats.

Hierarchical training of ViTs comprises multiple transformers being trained at different levels of resolution~\cite{si21,Chen22,chen2021crossvit, yu2023megabyte}. Training begins with the lowest-level transformer processing short sequence segments. Higher-level transformers iterate on using outputs from lower levels to process longer segments. However, employing multiple transformers increases the training time and memory usage. Moreover, managing multiple interacting transformers is complex, demanding hyperparameter tuning for the model at each resolution level.

\begin{table*}[t]
	\centering
	\footnotesize
	\setlength{\tabcolsep}{5pt}
 \caption{A summary of relevant long sequence training methods that reduce the amount of work. \emph{N} = sequence length.}
 \resizebox{\linewidth}{!}{
	\begin{tabular}{|c|l|c|l|c|c|}
		\hline
		\multicolumn{1}{|c|}{\tiny \textbf{Approach}} & \multicolumn{1}{c|}{\tiny \textbf{Method}} & \multicolumn{1}{c|}{\textbf{\tiny Merits \& Demerits} } & \multicolumn{1}{c|}{\tiny \textbf{Complexity (Best)} } & \multicolumn{1}{c|}{\tiny \textbf{Model}}
        & \multicolumn{1}{c|}{\tiny \textbf{Implementation}}\\ 

		\hline
		\multirow{8}{*}{\begin{tabular}[c]{@{}c@{}}\tiny Attention\\\tiny Approximation
            \end{tabular}}   
		&  \begin{tabular}[c]{@{}p{4cm}@{}}
                \tiny Longformer~\cite{beltagy2020longformer}\\\tiny ETC~\cite{ainslie2020etc}\\
            \end{tabular}    & \begin{tabular}[c]{@{}p{4cm}@{}}
                \tiny \textbf{(+)} Better time complexity vs Transformer. \\
                \tiny \textbf{(-)} 
                Sparsity levels insufficient for gains to materialize.
            \end{tabular}&
            \begin{tabular}[c]{@{}p{4cm}@{}}
                \tiny O$(N)$\\\tiny O$(N\sqrt{N})$
            \end{tabular}
                
            &\multirow{8}{*}{\begin{tabular}[c]{@{}c@{}}
                \tiny Some\\\tiny Models\\\tiny w/ Forked\\\tiny PyTorch
            \end{tabular}}
            &\multirow{8}{*}{\begin{tabular}[c]{@{}c@{}}
                \tiny Custom\\ \tiny Self-attention \\\tiny Implementation
            \end{tabular}}\\
           
            \cline{2-4}
            
		&   \begin{tabular}[c]{@{}l@{}}\tiny BigBird~\cite{zaheer2020big}\\\tiny Reformer~\cite{kitaev2020reformer}
            \end{tabular} & \begin{tabular}[c]{@{}p{4cm}@{}}
                \tiny \textbf{(+)} Theoretically proven time complexity.\\
                \tiny \textbf{(-)} High-order derivatives
            \end{tabular}& { \tiny O$(N\log{N})$}& &\\
		\cline{2-4}
            &  {\tiny Sparse Attention \cite{child2019generating}}  & \begin{tabular}[c]{@{}p{4cm}@{}}
                \tiny \textbf{(+)} Introduced sparse factorizations of the attention. \\
                \tiny \textbf{(-)} Higher time complexity.
            \end{tabular}& { \tiny O$(N\sqrt{N})$} & &\\
                \cline{2-4}
		& \begin{tabular}[c]{@{}c@{}}\tiny Linformer \cite{katharopoulos2020transformers}\\\tiny Performer \cite{choromanski2020rethinking}
            \end{tabular}
        & \begin{tabular}[c]{@{}p{4cm}@{}}
                \tiny \textbf{(+)} Fast adaptation \\
                \tiny \textbf{(-)} Assumption that self-attention is low rank.
            \end{tabular}& { \tiny O$(N)$}& &\\
		
            \hline
		\multirow{8}{*}{\tiny Hierarchical} 

		& {\begin{tabular}[c]{@{}p{4cm}@{}}\tiny Hier. Transformer~\cite{si21}\\\tiny (Text Classification)
            \end{tabular}}  & \begin{tabular}[c]{@{}p{4cm}@{}}
                \tiny \textbf{(+)} Independent hyperpara. tuning of hierarc. models.\\
                \tiny \textbf{(-)} No support for ViT.
            \end{tabular}& {\tiny O$(N\log{N}$)} & \multirow{8}{*}{\begin{tabular}[c]{@{}c@{}}
                \tiny Custom\\\tiny Model\\\tiny w/ Plain\\\tiny PyTorch
            \end{tabular}} 
            &\multirow{8}{*}{\begin{tabular}[c]{@{}c@{}}
                \tiny Custom \\\tiny Model \\\tiny Implementation
            \end{tabular}}\\
            \cline{2-4}
            		& {
\begin{tabular}[c]{@{}p{4cm}@{}}
                \tiny CrossViT~\cite{chen2021crossvit}\\\tiny (Classification)
            \end{tabular}}  & \begin{tabular}[c]{@{}p{4cm}@{}}
                \tiny \textbf{(+)} Better time complexity vs standard ViT. \\
                \tiny \textbf{(-)} Complex token fusion scheme in dual-branch ViTs.
            \end{tabular}& { \tiny O$(N)$}&  &\\
            \cline{2-4}
        & {\begin{tabular}[c]{@{}p{4cm}@{}}
                \tiny \tiny HIPT~\cite{Chen22}\\\tiny (Classification)
            \end{tabular}}  & \begin{tabular}[c]{@{}p{4cm}@{}}
                \tiny \textbf{(+)} Model inductive biases of features in the hierarchy. \\
                \tiny \textbf{(-)} High cost for training multiple models.
            \end{tabular}& { \tiny O$(N\log{N})$} & &\\
            \cline{2-4}
           & {\begin{tabular}[c]{@{}p{4cm}@{}}
                \tiny \tiny MEGABYTE~\cite{yu2023megabyte}\\\tiny (Prediction)
            \end{tabular}} & \begin{tabular}[c]{@{}p{4cm}@{}}
                \tiny \textbf{(+)} Support of multi-modality.\\
                \tiny \textbf{(-)} High cost for training multiple models.
            \end{tabular}& {\tiny O$(N^{\frac{4}{3}})$} & &\\
         
		\hline
		\multirow{1}{*}{{\textbf{\tiny Ours}}} 
		&  {\begin{tabular}[c]{@{}p{4cm}@{}}
                 \tiny \textbf{Adaptive Patching}\\\tiny \textbf{(Segmentation \& \tiny Class.)}
            \end{tabular}}          &\begin{tabular}[c]{@{}p{4cm}@{}}
                \tiny \textbf{(+)} Attention mechanism intact.   \\                
                \tiny \textbf{(+)} Negligible overhead. \\
                \tiny \textbf{(+)} Largely reduces computation cost; maintains quality. \\
                \tiny \textbf{(-)} Efficiency depends on level of details in an image. \\
            \end{tabular}& { \tiny O$(log^2N)$}&\begin{tabular}[c]{@{}c@{}}
                \tiny Any\\\tiny Model\\\tiny w/ Plain\\\tiny PyTorch
            \end{tabular} &\begin{tabular}[c]{@{}c@{}}
                \tiny Image\\\tiny Pre-processing
            \end{tabular}\\
		\hline
		\bottomrule[1.3pt]
    
        \end{tabular}
        }
	\label{tab:linear_ref}
\end{table*}


In summary, to scale long sequences for high-resolution image segmentation trained on ViT models or U-Net models that use transformers to ingest the images, we need the following: a) be able to use smaller patch sizes that are favorable in segmentation~\cite{strudel2021segmenter}, b) avoid the potential loss in performance that comes with self-attention altering mechanisms, c) avoid the high aggregate compute cost of sequence parallelism and tiling/blocking methods, and d) have a general solution that can work with transformer model, and not custom built models. 

To that end, we take inspiration from the tree-based Adaptive Mesh Refinement (AMR) methods pioneered~\cite{10.5555/891587} and used~\cite{10.1109/SC.2005.61,DBLP:conf/sc/WahibMA16} for decades in HPC to dramatically reduce the computational cost of solvers applied on structured discretized meshes. We propose an Adaptive Patch Framework (APF) that is compatible with any vision transformer. APF is a pre-processing solution that uses a quadtree to partition each image in the dataset into mixed-scale patches, based on the level of detail in different regions in the image. Larger patches, that carry fewer image details are then downscaled such that all patches become the same size when being fed to the model, while keeping the core attention mechanisms and ViT model architecture intact. To demonstrate APF's scalability, we conducted extensive training of transformer-based vision models with small patch sizes for long sequences of high-resolution images. The primary contributions outlined in this paper are as follows:


\begin{itemize}
   \item \textbf{Adaptive Patch Framework} A solution to reduce the total number of patches extracted from an image, thereby reducing the overall training cost. This not only reduces the cost of computing and memory, it also allows for using small sizes for patches, e.g., 4x4 or 2x2, which is favorable for high segmentation quality~\cite{strudel2021segmenter}. Our quantitative results demonstrate that at the same resolution levels $[512, 1024, 4096, 8192, 16384]$, a model using AFP can employ nearly $8\times$ smaller patch sizes or $64\times$ longer sequence lengths, while maintaining the same cost of traditional patching.
    
    \item \textbf{High-quality segmentation on real-world datasets} We conducted experiments on Frontier supercomputer, with up to $2,048$ MI250X using real-world high-resolution pathology datasets. At a fixed compute budget, and up to the depth of 13 multi-resolutions, we can scale to image resolutions up to $16K^2$, and lower the patch size from $16\times16$ to the minimum $2\times2$ on a vision transformer. Meanwhile, due to the smaller patch size at the same computational cost, we improve the segmentation quality by 5.5\% over widely used models. Alternatively, we can reach the same segmentation quality with speedups ranging between $12.7\times$ to $3.9\times$. We also demonstrate the versatility of AFP by achieving more than 7\% classification accuracy over the most sophisticated model for classifying of high-resolution microscopic pathology images.  

    \item \textbf{Simplicity and low-overhead} Unlike existing methods that modify attention mechanisms, our solution preserves the original attention mechanism. This ensures seamless integration into any vision transformer. APF is a very low-overhead pre-processing solution, that is further amortized over epochs: the overhead is effectively negligible.   
 
\end{itemize}

In summary, AFP offers a novel and general solution to the long-sequence challenge in ViTs, it preserves the dense self-attention merits, and reduces sequence length dramatically to boost the segmentation efficiency. This paves the way for enhanced applications of ViTs in high-resolution scientific imaging domains.

\section{Background and Motivation}

\subsection{Adaptive Mesh Refinement and Quadtrees in Imaging} 
Structured AMR~\cite{10.5555/891587} uses a hierarchical spatial representation of mesh spacing. In the 2D tree-based scheme, the mesh is organized into a hierarchy of refinement levels in a tree that represents the hierarchy of the mesh. The mesh is usually decomposed into relatively small fixed-sized quadrants of mesh cells. Each quadrant can be recursively refined into a set of quadrants of fine cells. A quadtree manages the mesh by maintaining explicit child-parent relationships between coarse and fine quadrants. At most one level of refinement difference is typically allowed between neighboring quadrants to maintain size relations. Traversing the quadrants across the three leaves corresponds to a Morton z-shaped space filling curve in the geometric domain~\cite{Tropf1981MultimensionalRS}. Accordingly, sorting the tree leaf blocks by their Morton ID would give a series of blocks that are affine in the geometric space of the mesh.

A similar concept appears in computer graphics, under the name of \emph{quadtrees}, where a mesh is replaced by an image, and mesh cells are replaced by image pixels. The history of quadtree structures dates back to early advancements in computer graphics and image processing~\cite{bergen1997efficient, klosowski1998efficient, bergen1997efficient, klosowski1998efficient, redding2007real, ericson2004real}. In the work of \cite{bergen1997efficient, klosowski1998efficient}, quadtrees (octrees) were used in 2D (3D) computer games to detect the collision of two objects efficiently in $O(n\log{n})$ time complexity, where $n$ is the number of particles. Quadtrees are also used as an image representation at different resolution levels and have been efficiently applied in image~\cite{bergen1997efficient} and video compression~\cite{klosowski1998efficient}. Recently, quadtrees have been used in image segmentation to improve attention efficiency, e.g., quadtree attention~\cite{tang2022quadtree}, and octree transformer~\cite{ibing2023octree}. Both of those approaches employ quadtrees, like the work in this paper. However, we introduce a quadtree-based pre-processing patching strategy without changing the model or attention scheme. In other words, our proposal doesn't involve additional complexity and custom model design; our solution can be integrated seamlessly into the current and future transformer-based encoders. 

\subsection{Vision Transformers and Attention}

Our proposed methods act as a pre-processing step to feed patches to vision transformers, or U-Net~\cite{ronneberger2015u} like models employing transformer encoders. ViTs~\cite{dosovitskiy2021image} comprise an embedding layer, transformer encoder layers, and a classification head. The embedding layer linearly projects the image patches sequence input into a sequence of flattened embeddings. Transformer encoder layers process these embeddings, capturing local and global context through self-attention mechanisms.

The attention mechanism in transformers computes attention scores $A$ between input tokens, forming the attention matrix. Let $x \in R^{N \times F}$ denote a sequence of $N$ feature vectors of dimensions $F$. A transformer is a function $T: R^{N \times F} \rightarrow R^{N \times F} $ defined by the composition of $L$ transformer layers $T_1(\cdot), . . . , T_L(\cdot)$ as follows,:
\begin{align}\label{eqn:trans}
T_l(x) = f_l(A_l(x) + x).
\end{align}
$A_l(\cdot)$ is the self-attention function. The function $f_l(\cdot)$ transforms each feature independently of the others, and is usually implemented with a small two-layer feedforward network. Formally, the input
sequence $x$ is projected by three matrices $W_Q \in R^{F \times D},W_K \in R^{F \times D}$, and $W_V \in R^{F \times D}$, to corresponding representations $Q$, $K$ and $V$. Thus, the attention scores are calculated as follows:
\begin{align}\label{eqn:atten}
Q&=xW_Q\\
K&=xW_K\\
V&=xW_V\\
A_{ij} &= \text{Softmax}\left(\frac{(Q_iK_j)^T}{\sqrt{d_k}}\right)
\end{align}
where $Q_i$ and $K_j$ are query and key vectors for tokens $i$ and $j$, and $d_k$ is the dimension of the key vectors. The complexity of the attention matrix is $O(N^2)$, where $N$ is the sequence length. The same is true for the memory requirements because the full attention matrix must be stored to compute the gradients for the weights of the queries, keys, and values. 

We further assume that the input is the content of a square image $x$ with a resolution of $Z$, that is, let $x \in R^{Z \times Z}$, and by assuming that patches arise from the uniform grid patch method of patch size $p$. Thus the sequence $N = (\frac{Z}{P})^2$. Therefore, the total computation and memory cost of attention scores according to resolution and patch size is $O([\frac{Z}{P}]^4)$. This complexity demonstrates the difficulties of increasing the resolution while decreasing patch size $P$ with the uniform grid patch strategy.  

\begin{figure*}[t]
\includegraphics[width=\textwidth]{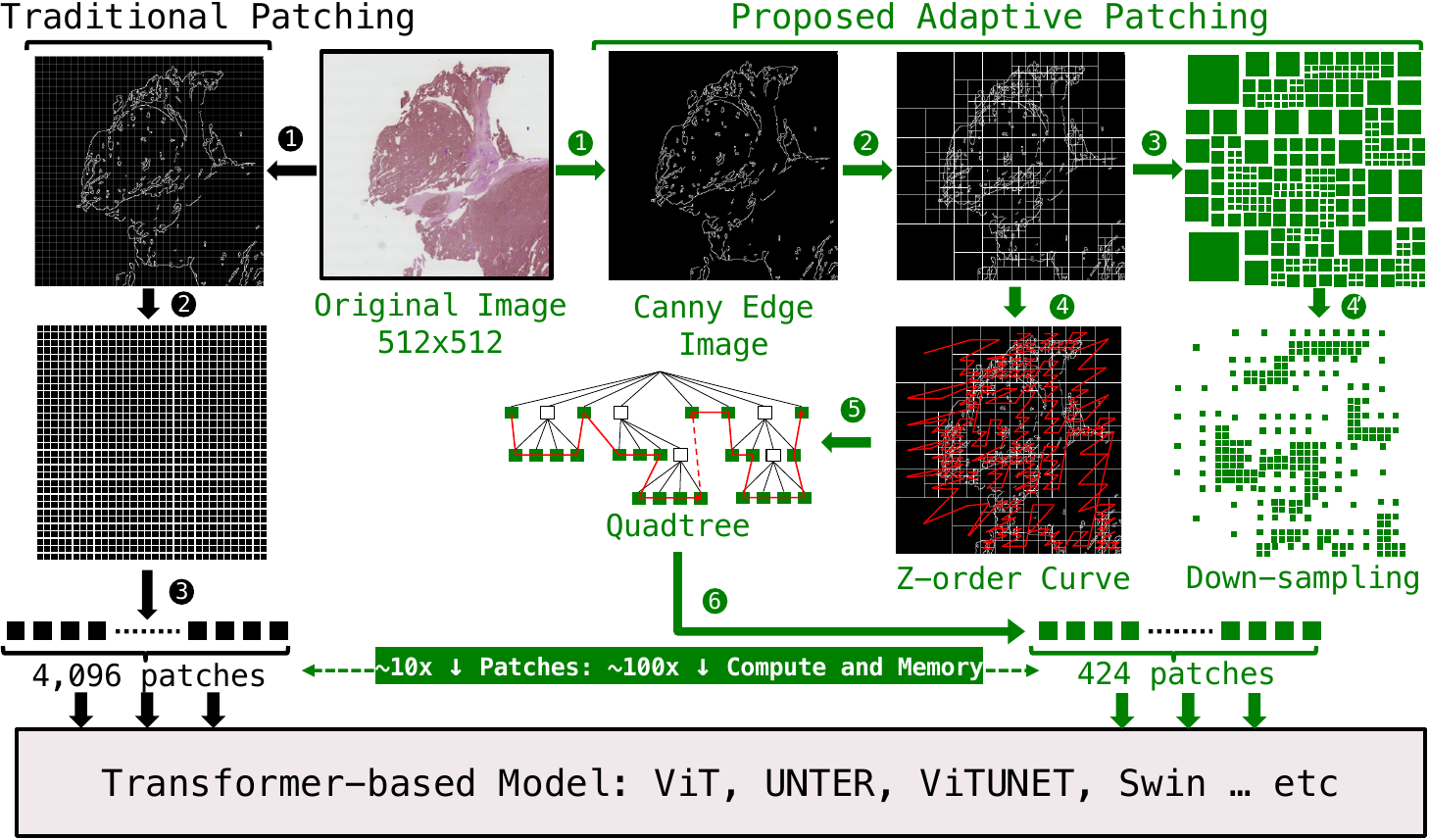}
    \caption{Overview of AFP. The right-side flow (green) shows all the steps, starting from the original image, and ending up with feeding the patches (tokens) to an intact transformer-based model. The reduction from 4,096 to 424 patches (of size $4\times4$) while achieving the same dice score is from a real example of training $512\times512$ images from the PAIP~\cite{KIM2021101854} liver cancer dataset on the UNTER~\cite{hatamizadeh2022unetr} model: $\sim9.6\times$ reduction in sequence length, and $\sim12.7\times$ speedup in end-to-end training.}
    \label{fig:quadtree_patchify}
\end{figure*}
\subsection{Long Sequence Problem}
Due to the quadratic cost of transformers w.r.t. the sequence length, numerous efforts have been dedicated to overcoming the long-sequence problem by reducing the amount of work. The first approach questions the necessity of full attention between all input embedding pairs. 
Longformer \cite{beltagy2020longformer} introduced a localized sliding window-based mask with few global masks to reduce computation scales linearly with the input sequence.  Child et al. \cite{child2019generating} proposed a set of sparse attention kernels that reduces the complexity to $O(n\sqrt{n})$ and saves memory usage of the backward pass. Reformer~\cite{kitaev2020reformer} further reduces the complexity to $O(n\log{n})$ based on locality-sensitive hashing. ETC \cite{ainslie2020etc} uses local and global attention instead of full self-attention to
scale transformers to long documents. BigBird \cite{zaheer2020big} is closely related to and built on the work of ETC. Linformer\cite{wang2020linformer} assumes the self-attention is low rank, and also proposes a linear complexity transformer. Later, Performers \cite{choromanski2020rethinking} also achieved linear space and time complexity and did not rely on any priors such as sparsity or low-rankness.

The second approach reduces the attention computation by training a hierarchy of models at different resolutions. Hierarchical transformers for text classification~\cite{si21} use three models to capture the structure in long sequences in documents. CrossViT~\cite{chen2021crossvit} classifies images by running a dual-attention model, in which each branch creates a non-patch token to exchange information with the other branch by attention. HIPT~\cite{Chen22} is a classifier for high-resolution images that trains multiple models at different resolutions to leverage the hierarchical geometric structure of visual tokens. The highest resolution model is trained with large patch sizes to reduce the sequence length. MEGABYTE~\cite{yu2023megabyte} predicts patches of bytes by running local and global models at different patch sizes.

We summarize the core idea of each approach and their limitations in Table~\ref{tab:linear_ref}. Hierarchical and attention approximation methods exploit the hierarchy and sparsity of the features inside the model. On the other hand, our solution is a lightweight mechanism that exploits the hierarchy and sparsity of features at different resolutions directly on the images in a pre-processing step, which leaves the attention mechanism and the model architecture intact.

\subsection{High-Resolution Segmentation} High Resolution (HR) aggravates the long-sequence problem. Initially, the common way in literature to handle this problem was to rely on a convolutional input encoder, which first down-samples the image to learn low-resolution features~\cite{chen2018encoder, long2015fully} and then up-sample to complete the prediction~\cite{badrinarayanan2017segnet}. To benefit from the effective entire-image receptive field of transformers, many efforts turned to transformer encoders (as pure ViT or CNN+ViT), and resorted to the techniques mentioned in the previous section for handling the long sequence problem. HRViT\cite{gu2022multi}, HRFormer\cite{yuan2021hrformer}, and HRNet\cite{wang2020deep} learn the HR representations by cross-resolution stream. 
Vision-LongFormer~\cite{zhang2021multi} uses a pyramid-like hierarchical structure of models at different scales to combine local attention and global memory. HIPT~\cite{Chen22} also applied a hierarchical pyramid transformer to a pathology dataset with the utmost $4K^2$ resolution. However, in comparison to these models, our method is a pre-processing strategy, which doesn't require additional revision to of the model or attention design. 

\section{Adaptive Patching for High-resolution Segmentation}
Figure~\ref{fig:quadtree_patchify} gives an overview of the flow of AFP, in comparison to the traditional method of dividing images uniformly into equal-sized patches. AFP divides the image into patches of different sizes based on the level of details, and then downsamples the large patches so that all patches have the same size. In the next section, we follow the flow of AFP starting from the original image up until the patches are fed to the model.  
\subsection{Quadtree-based Adaptive Patches}

\textbf{Image and Patches} We use the following notation to distinguish the size of an "image" and the "patch" corresponding to that image. Consider an image dataset $D$ consisting of input images $x \in R^{Z \times Z}$ where $Z$ is the resolution of image $x$.  Then, the sequence of non-overlapping patches can be noted as $\{x_i\}_{i=1}^{N} \in R^{N \times P}$ where $N$ is the sequence length and $P$ is the patch size. For the traditional uniform grid patching in ViT~\cite{dosovitskiy2021image}, the sequence length is $N=(\frac{Z}{P})^2$. For an image $x$ with resolution $Z=512$ (i.e. the image is $512\times512$) and patch size $P=8$ (i.e. the patch is $8\times8$), the sequence length $N$ is $4096$ patches (tokens). 


\textbf{Edge Extraction} To ignore the irrelevant details in the images $x$ in APF, as shown in step \circled{1} of Figure~\ref{fig:quadtree_patchify}, we apply Gaussian Blur with kernel $k$ and Canny~\cite{4767851} edge detection with lower $t_l$ and higher threshold $t_h$ to the original input images $x$. The Gaussian blur smooths the irrelevant details, and the Canny edge detection extracts the grayscale edges $x_e$ of the image. The kernel $k$ and threshold $t$ can also be used as hyper-parameters for controlling the smoothing effect. During our experiments, we kept the threshold as $[100, 200]$; the kernel size is set to be $[3,3,5,7,9,11,13]$ for resolutions $[512,1024,4096,8192,16384,32768,65536]$, respectively. 

\textbf{Quadtree Patches} The input edge $x_e$ undergoes a recursive quadtree partitioning shown by step \circled{2} of Figure~\ref{fig:quadtree_patchify}, creating nodes $Q_h$ that represent specific regions where $h$ is the depth of the quadtree. The quadtree node $Q_{h+1}$ is defined recursively as follows:
\begin{align}
Q_{h+1} = \begin{cases} 
Q_h & \text{if $\sum_i{D_i} \leq v$ or $h = H$ } \\
\{Q_{\text{NW}}^h, Q_{\text{NE}}^h, Q_{\text{SW}}^h, Q_{\text{SE}}^h\} & \text{if $\sum_i{D_i} > v$ and $h < H$}
\end{cases}
\end{align}
where $H$ is the maximum quadtree depth, $v$ is the subdivision criterion, $Q_{\text{NW}}^h, Q_{\text{NE}}^h, Q_{\text{SW}}^h, Q_{\text{SE}}^h$ are the $h$-th depth child nodes representing the northwest, northeast, southwest, and southeast quadrants, respectively~\cite{samet1984quadtree, finkel1974quad}. In our implementation, the subdivision criterion constraints the total number of pixels $\sum_i{D_i}$ confined in the data area by the split value $v$. The depth limitation $H$ is set to $[9,10,12,13,14,15,16]$ w.r.t. resolutions, which practically allows the input $x_e$ to be subdivided all the way down to the $2\times2$ patch size level. 

For uniform grid patches, we concatenate horizontal lines of patches into a 1D sequence of patches. On the other hand, for adaptive patching, after the quadtree is constructed, the patches, that is, the leaf nodes, need to be arranged. Here we show by steps \circled{4} and \circled{5} of Figure~\ref{fig:quadtree_patchify}, we use a Morton Z-order curve~\cite{Tropf1981MultimensionalRS} to arrange the nodes starting from the left end of the tree and going to the right. Z-order curves have the desirable property of keeping geometrically affine patches closer in the constructed sequence. After arranging the patches, since different images have different quadtree sequence lengths, firstly, we project all the different patches into the same minimized size $P_m$ (step \circled{4'} of Figure~\ref{fig:quadtree_patchify}). Next, we randomly drop or pad them to the same length $L$. Finally, the sequence of patches $x_p \in R^{L \times P_m} $ are fed to the model (step \circled{6} of Figure~\ref{fig:quadtree_patchify}) to train any underlying segmentation model using a transformer encoder $f(x_p;\theta)$.  We summarize the above steps in Algorithm~\ref{alg:apf}.

It is worth mentioning that for quadtree in the worst case, where all objects and details are in the same quadrant at the deepest level of the tree, the time complexity becomes $O(N^2)$. In the best cases, the quadtree patching strategy leads to $O(\log^2 N)$, where $N$ is the total number of patches. However, from empirical observations in pathology datasets, instead of the best or worst cases, we observed sub-linear growth in sequence length as the average patch size decreased. This linear complexity in sequence length suggests the empirical time complexity is approximately $O(n)$.


\begin{algorithm}[t]
\caption{Adaptive Patch Framework}\label{alg:apf}
\begin{algorithmic}[1]
\Require $v, H, k, t_l, t_h, f(x;\theta), N, T, D, D_p$

\State Initialize segment model $f(x;\theta)$.

\For{$n \xleftarrow{} 1$ \textbf{to} $N$}
\State $x_g = GaussianBlur(x_n ;k)$
\State $x_e = CannyEdge(x_g;(t_l, t_h))$
\State $x_p = QuadTreePatch(x_g; v, H)$
\State Add to $D_p = D_p \cup (x_p, x_n)$
\EndFor

\For{$t \xleftarrow{} 1$ \textbf{to} $T$}
\For{$n \xleftarrow{} 1$ \textbf{to} $N$}
    \State $x_p = D_p.pop()$
    \State Train $f(x;\theta)$ on the $x_p$. 
\EndFor
    \State Evaluate $f(x;\theta)$ on validation set.
\EndFor
\State Evaluate $f(x;\theta)$ on Test set.
\end{algorithmic}
\end{algorithm}






\section{Evaluation}
\subsection{Experimental Setup}
All the experiments were performed using the Frontier Supercomputer \cite{Frontier} at ORNL. Each Frontier node has a single 64-core AMD EPYC CPU and four AMD Instinct MI250X GPUs (128GB per GPU). The four MI250X GPUs are connected with Infinity Fabric GPU-GPU of 50GB/s. The nodes are connected via a Slingshot-11 interconnect with 100GB/s, to a total of 9,408 nodes. For the software stack, we used  Pytorch 2.4 nightly build 03/16/2024. ROCm v5.7.0, MIOpen v2.19.0, RCCL v2.13.4 with libfabric v1.15.2 plugin.
\begin{table*}[t]
	\centering
	\footnotesize
	\setlength{\tabcolsep}{5pt}
 	\caption{Speedup of AFP end-to-end training for PAIP dataset at the same segmentation quality of the baseline. We use the highest dice score of the baseline model (in Table~\ref{tab:quantitive_results}), and report the APF configuration with similar dice scores.}
	\begin{tabular}{c||l|r|c|c|c|c|c}
		\toprule[1.3pt]	
		\hline
        \multirow{2}{*}
		\thead{\textbf{Resolution}} & \thead{\textbf{Model-Patch}} & 
        \thead{\textbf{Sec/Image}}& \thead{\textbf{Sequence Length}} & \thead{\begin{tabular}[c]{@{}c@{}}
                \textbf{Quadtree}\\\textbf{Depth}
            \end{tabular}}&
        \thead{\begin{tabular}[c]{@{}c@{}}
                \textbf{Dice Score}\\(\%)
            \end{tabular}} & 
        \thead{\begin{tabular}[c]{@{}c@{}}
                \textbf{Speedup}\\(Sec/Image)
            \end{tabular}} 
            & 
        \thead{\begin{tabular}[c]{@{}c@{}}
                \textbf{Speedup}\\(Time to Convergence)
            \end{tabular}} \\
            \hline
            \multirow{2}{*}{ \begin{tabular}[c]{@{}c@{}}
                \shortstack{$512 \times 512$}\\ 1 GPU\\
            \end{tabular}    } 
		& \multirow{1}{*}{\textbf{APF-4}} & 
  0.06495& 1,024 & 7 & $77.88$   
            & \multirow{2}{*}{\shortstack{$7.48\times$}} & \multirow{2}{*}{\shortstack{$12.71\times$}} \\
            & \multirow{1}{*}{UNETR-4} & 
            0.4863& 16,384&-& $77.31$ &\\
  		\hline
            \multirow{2}{*}{\begin{tabular}[c]{@{}c@{}}
                \shortstack{$1,024 \times 1,024$}\\ 8 GPUs\\
            \end{tabular}    } 
		& \multirow{1}{*}{\textbf{APF-8}} & 
  0.14284 & 1,024& 7 & $75.63$
            & \multirow{2}{*}{\shortstack{$7.6\times$}}  & \multirow{2}{*}{\shortstack{$12.92\times$}}\\
            & \multirow{1}{*}{UNETR-8}  
            & 1.0863 & 16,384&- & $75.72$&\\
            \hline
            \multirow{2}{*}{\begin{tabular}[c]{@{}c@{}}
                \shortstack{$4,096 \times 4,096$}\\ 128 GPUs\\
            \end{tabular}    } 
		& \multirow{1}{*}{\textbf{APF-16}} 
  & 0.32231 & 2,116 & 8 & $75.74$
            & \multirow{2}{*}{\shortstack{$5.77\times$}}  & \multirow{2}{*}{\shortstack{$9.8\times$}}\\
            & \multirow{1}{*}{UNETR-32} & 
            1.8613 & 16,384&-& $75.77$ &\\
  		\hline
            \multirow{2}{*}{\begin{tabular}[c]{@{}c@{}}
                \shortstack{$8,192 \times 8,192$}\\ 256 GPUs\\
            \end{tabular}    } 
		& \multirow{1}{*}{\textbf{APF-16}} 
  & 1.1613 & 2,116 &9& $76.13$
            & \multirow{2}{*}{\shortstack{$2.29\times$}}  & \multirow{2}{*}{\shortstack{$3.89\times$}}\\
            & \multirow{1}{*}{UNETR-64} & 
            2.6618 & 16,384&-& $75.27$&\\
            \hline
            \multirow{2}{*}{\begin{tabular}[c]{@{}c@{}}
                \shortstack{$16,384 \times 16,384$}\\ 512 GPUs\\
            \end{tabular}    } 
		& \multirow{1}{*}{\textbf{APF-32}} 
  & 1.7613  & 1,024 &9& $75.92$
            & \multirow{2}{*}{\shortstack{$2.9\times$}} & \multirow{2}{*}{\shortstack{$4.93\times$}} \\
            & \multirow{1}{*}{UNETR-128}  
            & 5.1179 & 16,384&- & $75.89$ &\\
            \hline
            \multirow{2}{*}{\begin{tabular}[c]{@{}c@{}}
                \shortstack{$32,768 \times 32,768$}\\ 1024 GPUs\\
            \end{tabular}    } 
		& \multirow{1}{*}{\textbf{APF-32}} 
        & 2.1567 & 2,116&10 & $75.32$
            & \multirow{2}{*}{\shortstack{$3.79\times$}}  & \multirow{2}{*}{\shortstack{$6.44\times$}}\\
            & \multirow{1}{*}{UNETR-256} & 
            8.1896 & 16,384&- & $74.96$ &\\
  		\hline
            \multirow{2}{*}{\begin{tabular}[c]{@{}c@{}}
                \shortstack{$65,536 \times 65,536$}\\ 2048 GPUs\\
            \end{tabular}    } 
		& \multirow{1}{*}{\textbf{APF-32}} 
  & 5.733 & 4,096 &11& $75.82$
            & \multirow{2}{*}{\shortstack{$2.3\times$}}  & \multirow{2}{*}{\shortstack{$3.91\times$}}\\
            & \multirow{1}{*}{UNETR-512} & 
            13.218 & 16,384&- & $75.31$&\\
		\bottomrule[1.3pt]
	\end{tabular}
	\label{tab:speedup}
\end{table*}
\subsection{Datasets}
\textbf{PAIP}~\cite{KIM2021101854} is a high-resolution liver cancer pathology (real-world) dataset. The sample resolution size is close to $64K$, far higher than the resolution of conventional image datasets. PAIP includes $2,457$ Whole-Slide Images (WSIs). When needed to use smaller resolutions, we down-scale the images into uniform $[512, 1,024, 4,096, 8,192, 16,384, 32,768]$ square images. Before applying our quadtree patching method, we first apply Gaussian smoothing with kernel size $3\times3$ and $\sigma=0$. Then, we used Canny edge detection with a lower/higher threshold of $[100, 200]$ to extract the edges from the smoothed input. During the training process, we randomly select $0.7$ samples for training, $0.1$ samples for validation, and $0.2$ samples for testing. All data sets are shuffled and normalized to $[0.0,1.0]$ when used as model input.

\textbf{BTCV} challenge~\cite{landman2015miccai} for 3D multi-organ segmentation contains 30 subjects with abdominal CT scans where 13 organs are annotated by experts. Each CT scan consists of 80 to 225 slices with $512^2$ pixels. The multi-organ segmentation problem is formulated as a 13 classes segmentation task where the dice score typically reported is the average of the 13 classes. BTCV is relatively low in resolution in comparison to the PAIP dataset ($512^2$ vs. $64K^2$), yet is widely used as a benchmark by the high-resolution medical segmentation community. 

\subsection{Models}
Because our method is a patching strategy, it can easily replace the uniform grid patching method typically used in transformers. In our experiments, we use one of the widely-used models, UNETR~\cite{hatamizadeh2022unetr}, as the baseline model we use for AFP to conduct experiments on the high-resolution medical image segmentation task. It is worth nothing that in all our results we train the model from scratch for the target dataset: we do not do any pre-training on other datasets or fine-tune. We also report results for various other highly performing models as baselines, TransUnet~\cite{DBLP:journals/corr/abs-2102-04306}, HIPT~\cite{Chen22}, Swin UNETR~\cite{DBLP:conf/cvpr/TangY0RLXNH22}, ViT~\cite{dosovitskiy2021image}, and U-Net~\cite{ronneberger2015u}, to demonstrate different aspect about the performance and efficiency of AFP. 

UNETR uses a contraction-expansion pattern consisting of several transformers as an encoder. It is connected to the decoder via a skip connection. UNETR's initial target application was 3D medical imaging for human organs. The original work~\cite{hatamizadeh2022unetr} also discussed the impact of patch size on the model: the smaller the patch size, the better the model performance will be. However, due to the memory capacity and compute power limitation associated with quadratic attention, the authors reported that conducting experiments with a small patch size is unfeasible.
Since our target experimental data is 2D medical images, we only swap the 3D convolution and deconvolution blocks in UNETR with the 2D version without additional changes to the model structure. Other than that, we make no changes nor do we tune the original UNETR model.

\subsection{Training Setup}

The loss function we applied is a combination of dice loss and binary cross-entropy loss:
\begin{align}\label{eqn:dice_bce}
L(\hat{y}, y)& = w \cdot L_{bce}(\hat{y}, y) + (1 - w) \cdot L_{dice}(\hat{y}, y) \\
    &= - w \cdot \frac{1}{N} \sum_{i=1}^{N} [y_i \log(\hat{y}_i) + (1 - y_i) \log(1 - \hat{y}_i)] \\
    &+ (1 - w) \cdot(1 - \frac{2 \sum_{i=1}^{N} (\hat{y}_i \cdot y_i) + \epsilon}{\sum_{i=1}^{N} \hat{y}_i + \sum_{i=1}^{N} y_i + \epsilon} )
\end{align}
where $L(\hat{y}, y)$ represents the combined loss function, composed of a weighted sum of Binary Cross-Entropy (BCE) loss and dice loss. $w$ is the weight parameter controlling the contribution of BCE loss versus the dice loss:, we set it to 0.5 during the experiments. $\epsilon$ is a smoothing term, and we keep it to $1.0$ during the experiments. For the resolutions $[512, 1024, 4,096]$, all models were trained with a batch size of $16$, using the AdamW optimizer~\cite{loshchilov2017decoupled} with an initial learning rate of 0.0001 for $300$ epochs and decay by a factor of 0.1 at epoch step $[500,750,875]$. For the resolutions $[8,192, 16,384, 32,768, 65,536]$, we countered the problem of fitting a single sample in memory by tuning the sequence length and training for $200$ epochs.
\begin{table*}[t!]
	\centering
	\footnotesize
	\setlength{\tabcolsep}{5pt}
 	\caption{Improvement in quality of segmentation for the PAIP dataset against different baselines.}
	\begin{tabular}{c||l|c|c|r|c|r|c|c}
		\toprule[1.3pt]	
		\hline
		\thead{\textbf{Resolution}} & {\textbf{Model}} & \thead{\textbf{Patch}} & \thead{\textbf{GPUs}}& \thead{\textbf{Sec/Image/GPU}}& \thead{\textbf{Depth}}& \thead{\textbf{Sequence Length}} & \thead{\textbf{Dice Score}} & \thead{\textbf{Dice Improvement}} \\
		\hline
		\multirow{8}{*}{\shortstack{$512 \times 512$}} 
            & \multirow{3}{*}{\begin{tabular}[l]{@{}l@{}}
            \textbf{\shortstack{APF}}\\\textbf{\shortstack{(+UNTER)}}
            \end{tabular}}
		& \multirow{1}{*}{2} & 
  1& 0.06112 & 8 & 729 & \textbf{78.32 
  }& \multirow{9}{*}{4.11\%}\\
		\cline{3-8}
		&& \multirow{1}{*}{4} & 
  1& 0.05975 &7 & 676 & $77.88$&\\
            \cline{3-8}
  		&& \multirow{1}{*}{8} & 
  1& 0.05812 &6 & 576 & $75.17$&\\
            \cline{2-8}
            & \multirow{3}{*}{\shortstack{UNETR}}
            & \multirow{1}{*}{4} & 
  1& 0.4863 & - & 16,384 &   77.31&\\
            \cline{3-8}
		&& \multirow{1}{*}{8} & 
  1& 0.3746 &- & 4,096& $75.23$&\\
            \cline{3-8}
		&& \multirow{1}{*}{16} & 
  1& 0.1477 &- & 1,024& 74.88&\\
		\cline{2-8}
            & \multirow{1}{*}{\shortstack{TransUNet}}
            & \multirow{1}{*}{-} & 
  1& 0.1783 &- & 1,024 & 73.32&\\
            \cline{2-8}
            & \multirow{1}{*}{\shortstack{U-Net}}
            & \multirow{1}{*}{-} & 
  1& 0.0438 &- & - & 70.32&\\
            \hline
  		\multirow{8}{*}{\shortstack{$1,024 \times 1,024$}} 
             & \multirow{4}{*}{\begin{tabular}[l]{@{}l@{}}
            \textbf{\shortstack{APF}}\\\textbf{\shortstack{(+UNTER)}}
            \end{tabular}}
    	& \multirow{1}{*}{2} & 
  8& 0.2314 & 9  & 1,024 & \textbf{78.42 
     } & \multirow{9}{*}{7.10\%}\\ 
            \cline{3-8}
		&& \multirow{1}{*}{4} & 
  8& 0.1786 &8 & 900 & $77.64$\\
            \cline{3-8}
		&& \multirow{1}{*}{8} & 
  8& 0.1428 &7 & 784 & $75.63$\\
		\cline{3-8}
		&& \multirow{1}{*}{16} & 
  8& 0.1313 &6  & 576 & $74.88$\\
		\cline{2-8}
            & \multirow{4}{*}{\shortstack{UNETR}}
		& \multirow{1}{*}{8} & 
  32& 1.0863 & -  & 16,384 & 75.72\\
		\cline{3-8}
		&& \multirow{1}{*}{16} & 
  16& 0.9731 & -  & 4,096 & 75.12\\
		\cline{3-8}
		&& \multirow{1}{*}{32} & 
   8& 0.8874 & -  & 1,024 & 73.22\\
  		\cline{2-8}
            & \multirow{1}{*}{\shortstack{TransUNet}}
            & \multirow{1}{*}{-} & 
   8& 1.3247 & - & 4,096 &  72.38\\
            \cline{2-8}
            & \multirow{1}{*}{\shortstack{U-Net}}
            & \multirow{1}{*}{-} & 
   1& 0.0981 & - & - & 68.92\\
            \hline
            \multirow{6}{*}{\shortstack{$4,096 \times 4,096$}} 
            & \multirow{4}{*}{\begin{tabular}[l]{@{}l@{}}
            \textbf{\shortstack{APF}}\\\textbf{\shortstack{(+UNTER)}}
            \end{tabular}}
		& \multirow{1}{*}{2} & 
  128&0.6938& 11 & 4,096 & \textbf{79.63 
  } & \multirow{7}{*}{5.09\%}\\
		\cline{3-8}
		&& \multirow{1}{*}{4} & 
  128&0.4695& 10  & 2,116 & 78.17\\
		\cline{3-8}
		&& \multirow{1}{*}{8} & 
  64 &0.3824&9 & 1,521 & 75.74\\
		\cline{3-8}
		&& \multirow{1}{*}{16}& 
  32 & 0.3223 &8 & 1,024 & 74.96\\
  		\cline{2-8}
            & \multirow{1}{*}{\shortstack{UNETR}}
		& \multirow{1}{*}{32} & 
  128 &1.8613& - & 16,384 & 75.77\\
    	\cline{2-8}
            & \multirow{1}{*}{\shortstack{TransUNet}}
            & \multirow{1}{*}{-} & 
  128 & 2.1637 & - & - & 71.32\\
            \cline{2-8}
            & \multirow{1}{*}{\shortstack{U-Net}}
            & \multirow{1}{*}{-} & 
   16 & 0.3712& - & - & 64.11\\
            \hline
            \multirow{6}{*}{\shortstack{$8,192 \times 8,192$}}
            & \multirow{4}{*}{\begin{tabular}[c]{@{}l@{}}
            \textbf{\shortstack{APF}}\\\textbf{\shortstack{(+UNTER)}}
            \end{tabular}}
		& \multirow{1}{*}{2} & 
  256 & 2.3314& 12 & 10,609 & \textbf{79.56 
  } & \multirow{7}{*}{5.70\%}\\
		\cline{3-8}
		&& \multirow{1}{*}{4} & 
  256 & 2.1314& 11  & 8,464 & 78.31\\
		\cline{3-8}
		&& \multirow{1}{*}{8} & 
  128 & 1.7867 & 10  & 4,096 & 77.61\\
		\cline{3-8}
		&& \multirow{1}{*}{16} & 
  64 & 1.1613& 9 & 2,116 & 76.13\\
  		\cline{2-8}
            & \multirow{1}{*}{\shortstack{UNETR}}
		& \multirow{1}{*}{64} &
  256 &2.6618& -  & 16,384 &75.27\\
    	\cline{2-8}
            & \multirow{1}{*}{\shortstack{TransUNet}}
            & \multirow{1}{*}{-} & 
  256 & 2.3678 & - & - & 70.89\\
            \cline{2-8}
            & \multirow{1}{*}{\shortstack{U-Net}}
            & \multirow{1}{*}{-} & 
   32 & 1.2858 & - & - & 63.21\\
		\hline
            \multirow{6}{*}{\shortstack{$16,384 \times 16,384$}}
            & \multirow{4}{*}{\begin{tabular}[c]{@{}l@{}}
            \textbf{\shortstack{APF}}\\\textbf{\shortstack{(+UNTER)}}
            \end{tabular}}
		& \multirow{1}{*}{2} & 
  512 & 4.8792& 13 & 16,384 & \textbf{80.62 
  } & \multirow{7}{*}{6.23\%}\\
		\cline{3-8}
		&& \multirow{1}{*}{4} & 
  256& 3.1231& 12  & 8,464 & 79.31\\
		\cline{3-8}
		&& \multirow{1}{*}{8} & 
  256& 1.8574& 11  & 4,096 & 78.84\\
		\cline{3-8}
		&& \multirow{1}{*}{16} & 
  128&1.6421& 10 & 2,116 & 77.43\\
		\cline{2-8}
            & \multirow{1}{*}{\shortstack{UNETR}}
		& \multirow{1}{*}{128} & 
  512 & 5.1179& -  & 16,384 & 75.89 \\
      	\cline{2-8}
            & \multirow{1}{*}{\shortstack{TransUNet}}
            & \multirow{1}{*}{-} & 
 512&  6.1296& - & - & 70.46\\
            \cline{2-8}
            & \multirow{1}{*}{\shortstack{U-Net}}
            & \multirow{1}{*}{-} & 
 256& 2.7825 & - & - & 62.97\\
		\hline
            \multirow{6}{*}{\shortstack{$32,768 \times 32,768$}}
            & \multirow{4}{*}{\begin{tabular}[c]{@{}l@{}}
            \textbf{\shortstack{APF}}\\\textbf{\shortstack{(+UNTER)}}
            \end{tabular}}
		& \multirow{1}{*}{4} & 
  1024 &7.8916& 13 & 16,384 & \textbf{78.98 
  } & \multirow{7}{*}{5.36\%}\\
		\cline{3-8}
		&& \multirow{1}{*}{8} & 
  512 & 6.1792 & 12  & 8,464 & 78.31\\
		\cline{3-8}
		&& \multirow{1}{*}{16} & 
  512 & 4.1685 & 11  & 4,096 & 77.61\\
		\cline{3-8}
		&& \multirow{1}{*}{32} & 
  256 & 2.1567& 10 & 2,116 & 76.13\\
  		\cline{2-8}
            & \multirow{1}{*}{\shortstack{UNETR}}
		& \multirow{1}{*}{256} & 
  1024 & 8.1896 & - & 16,384 & 74.96\\
  		\cline{2-8}
            & \multirow{1}{*}{\shortstack{TransUNet}}
            & \multirow{1}{*}{-} & 
  1024 & 10.001 & - & - & 69.88\\
            \cline{2-8}
            & \multirow{1}{*}{\shortstack{U-Net}}
            & \multirow{1}{*}{-} & 
  512 & 4.2714 & - & - & 61.38\\
        \hline
   \multirow{6}{*}{\shortstack{$65,536 \times 65,536$}} 
            & \multirow{4}{*}{\begin{tabular}[c]{@{}l@{}}
            \textbf{\shortstack{APF}}\\\textbf{\shortstack{(+UNTER)}}
            \end{tabular}}
		& \multirow{1}{*}{8} & 
  2048 & 12.697 & 13 & 16,384 & \textbf{77.77 
  } & \multirow{7}{*}{3.27\%}\\
		\cline{3-8}
		&& \multirow{1}{*}{16} & 
  1024 & 8.793 & 12  & 8,464 & 76.11\\
		\cline{3-8}
		&& \multirow{1}{*}{32} & 
  512 & 5.733 & 11  & 4,096 & 75.41\\
		\cline{3-8}
		&& \multirow{1}{*}{64} & 
  256 & 3.961 & 10 & 2,116 & 75.13\\
		\cline{2-8}
            & \multirow{1}{*}{\shortstack{UNETR}}
		& \multirow{1}{*}{512} & 
  2048 & 13.218 & -  & 16,384 & 75.31\\
    	\cline{2-8}
            & \multirow{1}{*}{\shortstack{TransUNet}}
            & \multirow{1}{*}{-} & 
  2048 & 14.3516 & - & - & 67.67\\
            \cline{2-8}
            & \multirow{1}{*}{\shortstack{U-Net}}
            & \multirow{1}{*}{-} & 
  1024 & 5.961 & - & - & 59.69\\
		\hline
		\bottomrule[1.3pt]
	\end{tabular}
	\label{tab:quantitive_results}
\end{table*}
\subsection{Evaluation Metrics}
For computational performance, we report the seconds/image of end-to-end training. For the quantitative evaluation of the segmentation result, we use the dice score, which measures the similarity between a predicted segmentation mask and the ground truth segmentation mask. The dice score (also known as the dice similarity coefficient) is defined as:
\begin{equation*}
\text{Dice(X,Y)} = \frac{2 \times |X \cap Y|}{|X| + |Y|}
\end{equation*}
where $X$ and $Y$ are the two sets being compared. $|X \cap Y|$ represents the cardinality of the intersection of sets $X$ and $Y$. $|X|$ and $|Y|$ represent the cardinality of sets $X$ and $Y$ respectively. A dice score of 100\% means identical similarity between the prediction and the ground truth.


\subsection{Results}
\subsubsection{\textbf{Speedup of End-to-end Training at the Same Segmentation Quality}}
In Table~\ref{tab:speedup} we show that under the same dice score, AFP is just a pre-processing step (on top of UNTER as baseline) that achieves a geomean speedup of $4.1\times$, if we compare on the basis that both AFP and the baseline run to the same number of epochs. Since we further observe the convergence speed in AFP to be $1.7\times$ faster, the speedup to get to the same dice score goes up to the geomean speedup of $6.9\times$. At the highest resolution of $64^2$ training on 2,048 GPUs, AFP achieves $\sim$4$\times$ speedup. It is worth mentioning that AFP also brings significant savings in memory and not just speedup.  

\subsubsection{\textbf{Gain in Segmentation Quality}}
Table~\ref{tab:quantitive_results} shows segmentation improvement over different models, at different PAIP resolutions. At similar resolution, with adaptive patches we can use nearly $8\times$ smaller patch sizes at the same, computational complexity, and improve upon the original model dice score with an average of $5.5\%$. It is worth noting that on top of improving the dice score, we achieve those improvements with additional speedups to the training time up to $4.6\times$. 

Table~\ref{table:btcv_result} shows segmentation results for BTCV ($512^2$ resolution). Following~\cite{zhou2017fixed, chen2021transunet}, we applied APF to each 2D slice of each CT sample and inferred all the slices to reconstruct the final 3D predictions. As shown in the table, AFP-UNTER gives higher quality than other models, with the exception of Swin UNTER (which has the advantage of being pre-trained on five other datasets before fine-tuning on BTCV). On top of getting the highest dice score, this is achieved at $>$8$\times$ faster training time over models with similar dice score. 
\begin{table}[t]
	
    	\caption{Segmentation of BTCV~\cite{landman2015miccai} for multi-organ segmentation on one GPU. \emph{Time} reported is the end-to-end time to reach the reported dice score.}
    \centering
    \resizebox{\linewidth}{!}{
	\begin{tabular}{l||c|r|c|c}
		\toprule[1.3pt]	
		\hline
		\thead{\textbf{Model}} & \thead{\textbf{Patch Size}} & \thead{\textbf{Time}}  & \thead{\textbf{Speedup}} & \thead{\textbf{Dice (\%)}} \\
            \hline
            U-Net~\cite{ronneberger2015u} & N/A & 843.90 Sec    & 0.79$\times$ &  80.2  \\
            \hline
            TransUNet~\cite{DBLP:journals/corr/abs-2102-04306} & N/A & 3115.25 Sec  & 2.91$\times$& 83.8 \\
             \hline
            UNETR~\cite{hatamizadeh2022unetr} &  4 & 8386.56 Sec  & 7.85$\times$ &  89.1  \\
             \hline
            Swin UNETR$^{*}$~\cite{DBLP:conf/cvpr/TangY0RLXNH22} & 4 & 6609.45 Sec  & 6.19$\times$ &  91.8  \\
             \hline
            \textbf{APF-UNETR} & 2 & 1067.88 Sec  & 1$\times$ &  89.7  \\
		\hline
  \multicolumn{4}{l}{$^{*}$Unlike APF-UNTER, Swin UNTER is pre-trained on five datasets.}
	\end{tabular}}
	\label{table:btcv_result}
\end{table}
\subsubsection{\textbf{Segmentation Qualitative Results}}
\begin{figure*}[t!]
    \begin{subfigure}[b]{0.18\textwidth}
        \caption{$512^2@100\%$}
        \includegraphics[width=\textwidth]{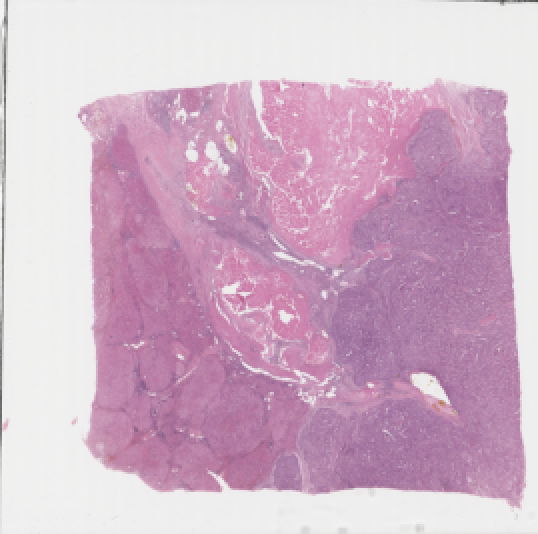}
    \end{subfigure}
    \hfill
    \begin{subfigure}[b]{0.18\textwidth}
        \caption{Dice Score:100\%}
        \includegraphics[width=\textwidth]{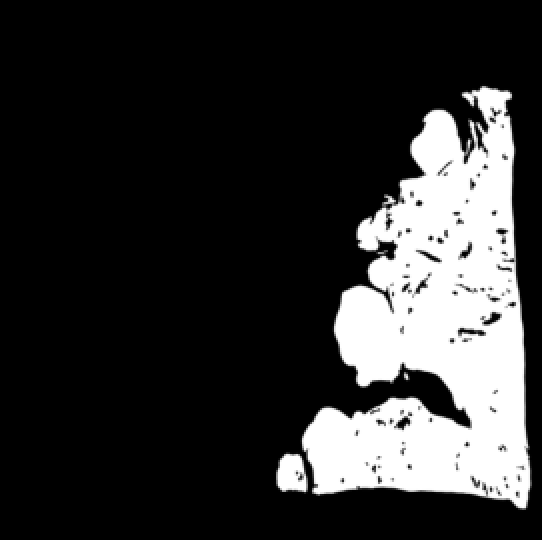}
    \end{subfigure}
        \hfill
    \begin{subfigure}[b]{0.18\textwidth}
        \caption{73.32\%}
        \includegraphics[width=\textwidth]{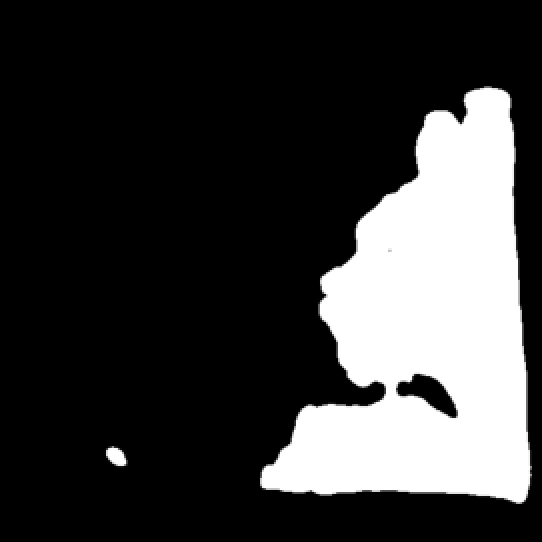}
    \end{subfigure}
        \hfill
    \begin{subfigure}[b]{0.18\textwidth}
        \caption{77.31\%}
        \includegraphics[width=\textwidth]{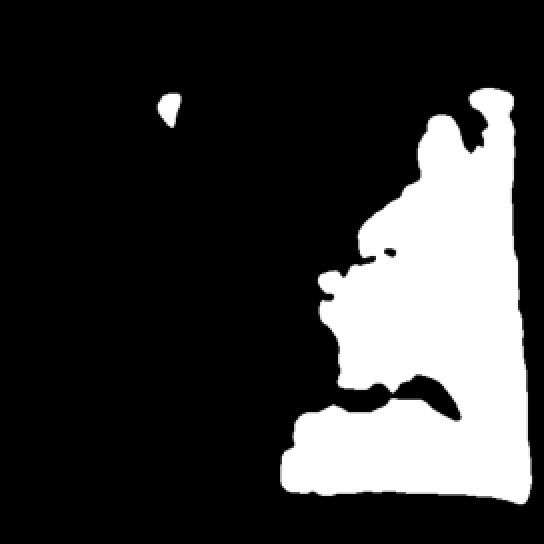}
    \end{subfigure}
    \hfill
    \begin{subfigure}[b]{0.18\textwidth}
        \caption{78.32\%}
        \includegraphics[width=\textwidth]{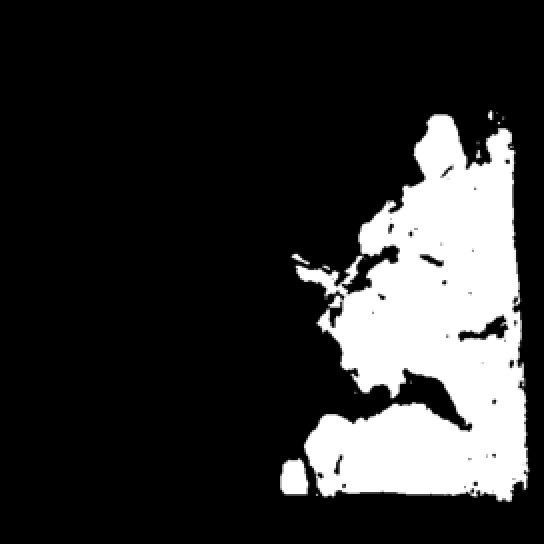}
    \end{subfigure}
    \hfill
    \begin{subfigure}[b]{0.18\textwidth}
        \caption{$4,096^2@1.5\%$}
        \includegraphics[width=\textwidth]{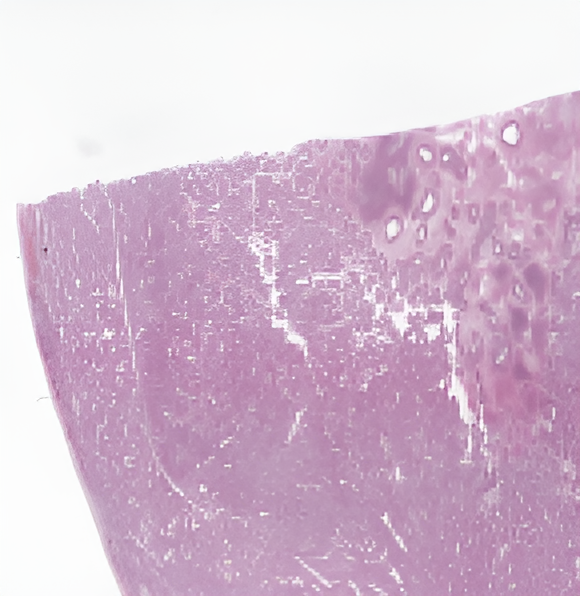}
    \end{subfigure}
    \hfill
    \begin{subfigure}[b]{0.18\textwidth}
        \caption{Dice Score:100\%}
        \includegraphics[width=\textwidth]{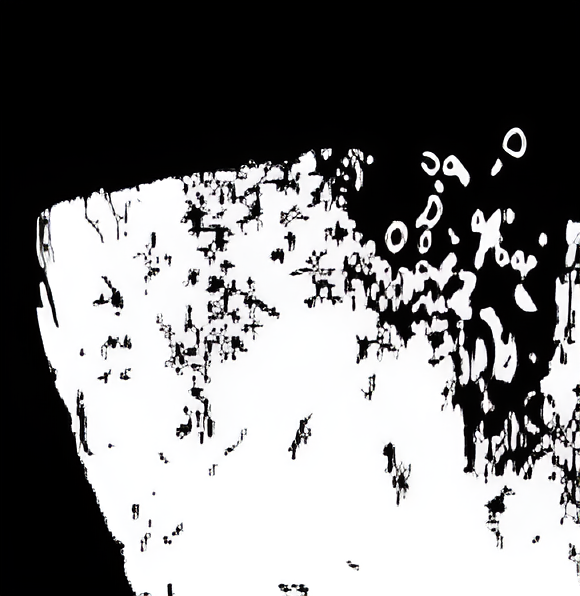}
    \end{subfigure}
        \hfill
    \begin{subfigure}[b]{0.18\textwidth}
        \caption{71.32\%}
        \includegraphics[width=\textwidth]{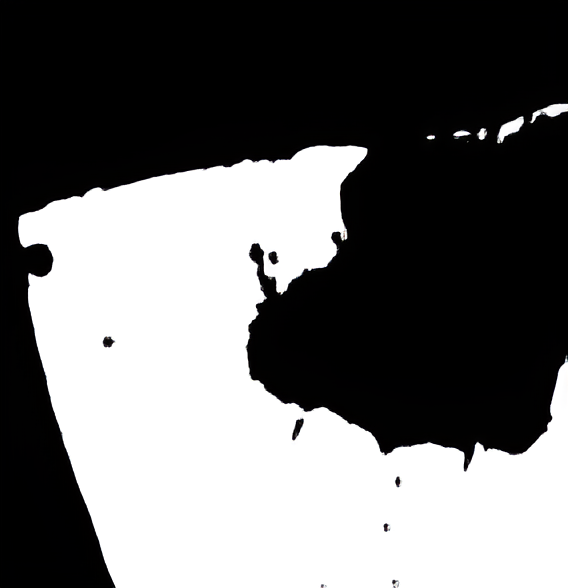}
    \end{subfigure}
        \hfill
    \begin{subfigure}[b]{0.18\textwidth}
        \caption{75.77\%}
        \includegraphics[width=\textwidth]{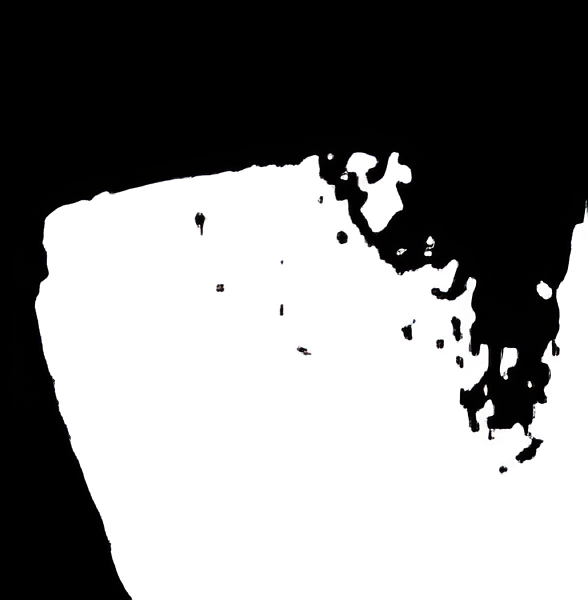}
    \end{subfigure}
    \hfill
    \begin{subfigure}[b]{0.18\textwidth}
        \caption{79.63\%}
        \includegraphics[width=\textwidth]{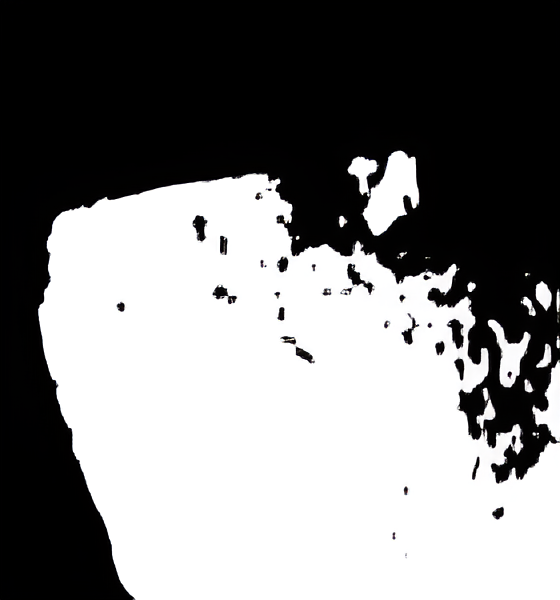}
    \end{subfigure}
    \hfill
    \begin{subfigure}[b]{0.18\textwidth}
        \caption{$8,192^2@0.39\%$}
        \includegraphics[width=\textwidth,height=3.4cm]{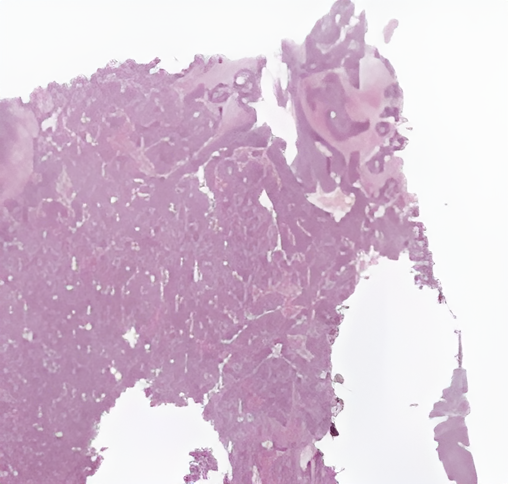}
    \end{subfigure}
    \hfill
    \begin{subfigure}[b]{0.18\textwidth}
        \caption{Dice Score:100\%}
        \includegraphics[width=\textwidth,height=3.4cm]{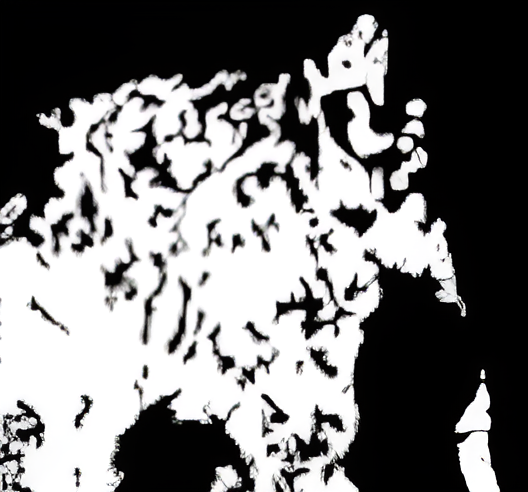}
    \end{subfigure}
        \hfill
    \begin{subfigure}[b]{0.18\textwidth}
        \caption{71.32\%}
        \includegraphics[width=\textwidth,height=3.4cm]{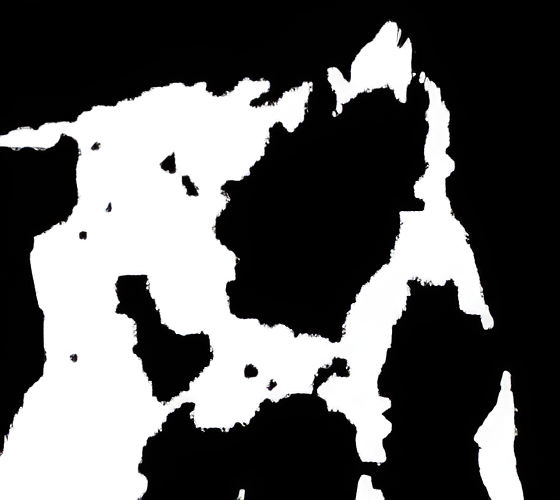}
    \end{subfigure}
        \hfill
    \begin{subfigure}[b]{0.18\textwidth}
        \caption{75.77\%}
        \includegraphics[width=\textwidth,height=3.4cm]{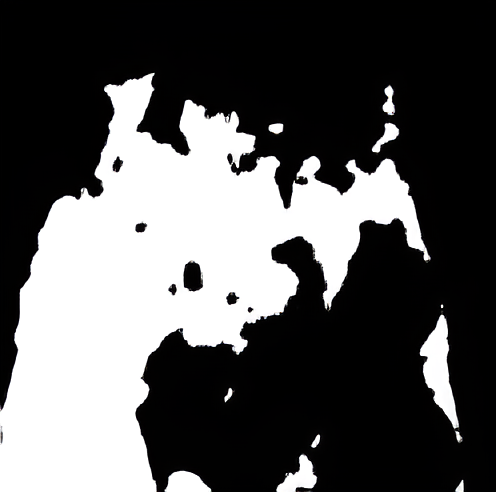}
    \end{subfigure}
    \hfill
    \begin{subfigure}[b]{0.18\textwidth}    
         \caption{79.63\%}
        \includegraphics[width=\textwidth,height=3.4cm]{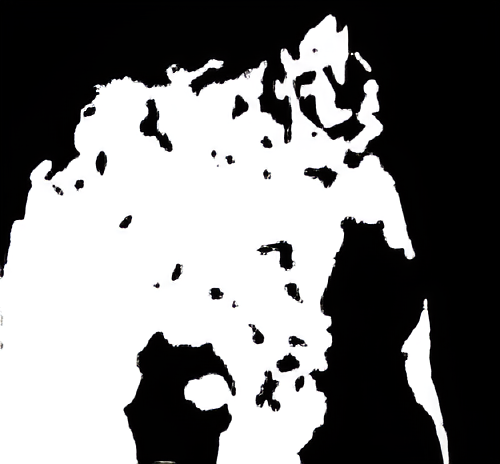}
    \end{subfigure}
    \hfill
    \begin{subfigure}[b]{0.18\textwidth}
        \caption{$32,768^2@0.024\%$}
        \includegraphics[width=\textwidth,height=3.4cm]{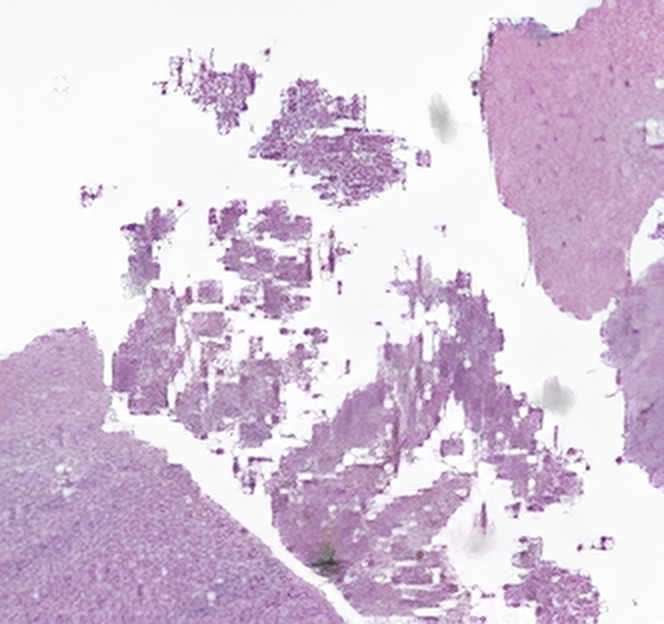}
    \end{subfigure}
    \hfill
    \begin{subfigure}[b]{0.18\textwidth}
        \caption{Dice Score:100\%}
        \includegraphics[width=\textwidth,height=3.4cm]{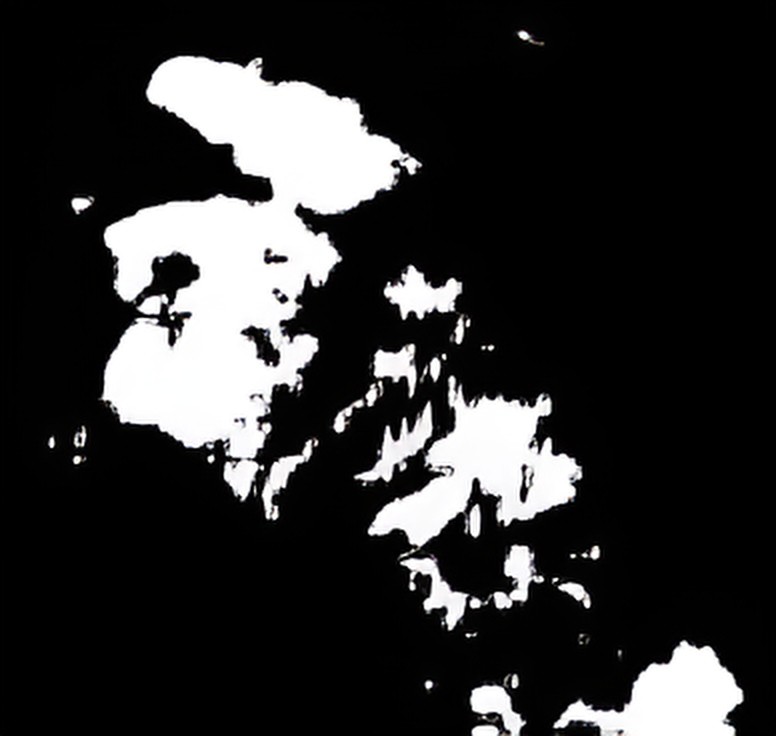}
    \end{subfigure}
        \hfill
    \begin{subfigure}[b]{0.18\textwidth}
        \caption{69.88\%}
        \includegraphics[width=\textwidth,height=3.4cm]{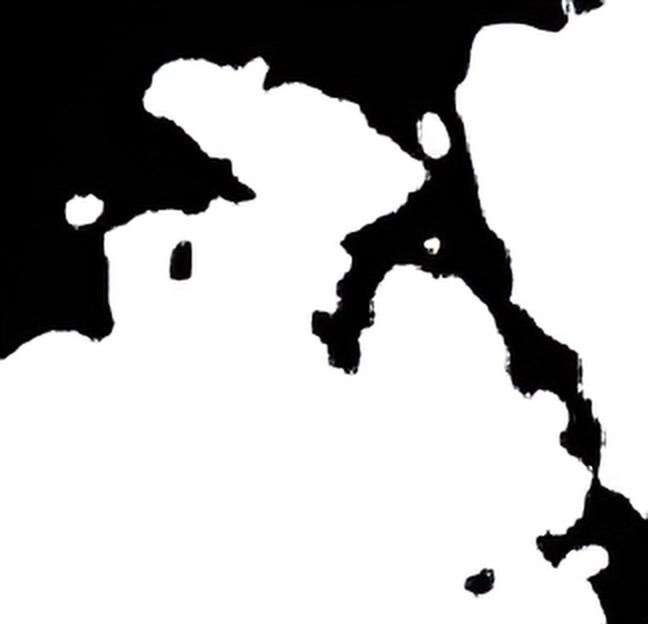}
    \end{subfigure}
        \hfill
    \begin{subfigure}[b]{0.18\textwidth}
        \caption{74.96\%}
        \includegraphics[width=\textwidth,height=3.4cm]{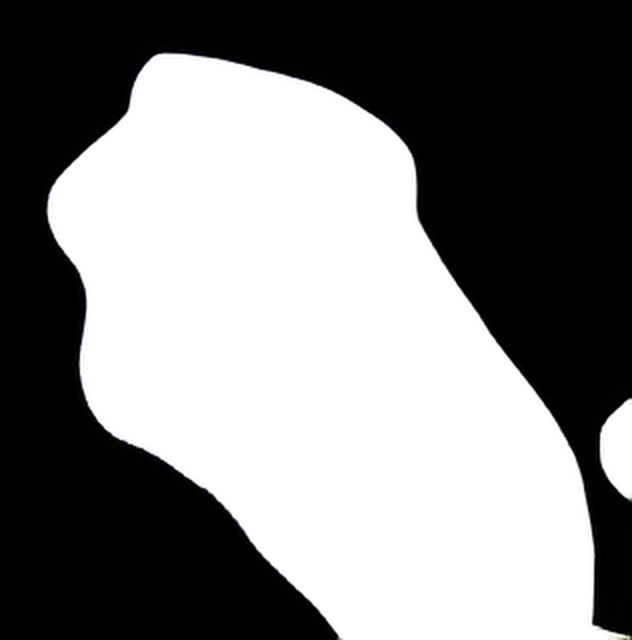}
    \end{subfigure}
    \hfill
    \begin{subfigure}[b]{0.18\textwidth}
        \caption{78.98\%}
        \includegraphics[width=\textwidth,height=3.4cm]{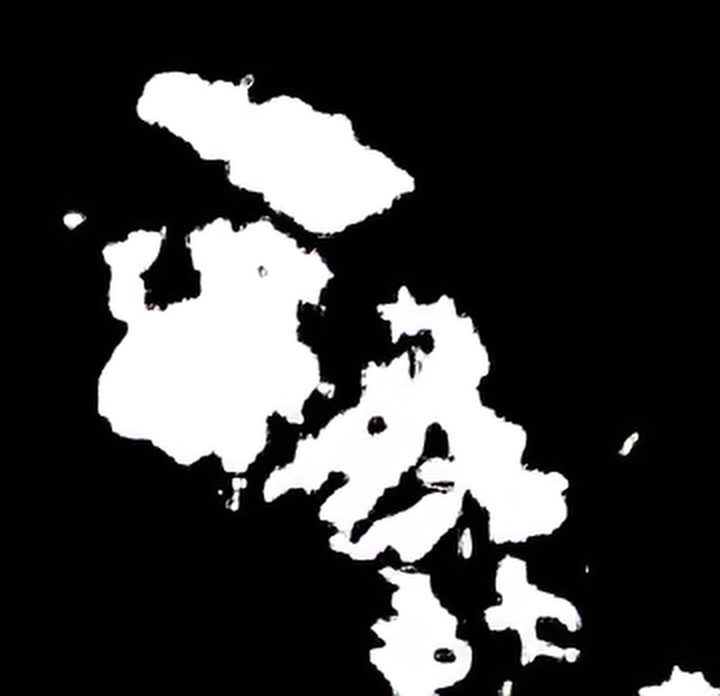}
    \end{subfigure}
    \hfill
    \begin{subfigure}[b]{0.18\textwidth}
        \caption{$65,5346^2@0.006\%$}
        \includegraphics[width=\textwidth,height=3.4cm]{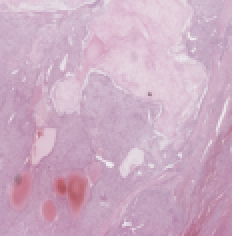}
        \caption{PAIP dataset images}
    \end{subfigure}
    \hfill
    \begin{subfigure}[b]{0.18\textwidth}
        \caption{Dice Score:100\%}
        \includegraphics[width=\textwidth,height=3.4cm]{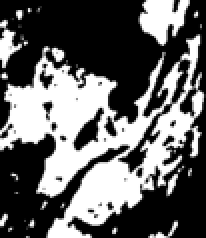}
        \caption{Ground Truth}
    \end{subfigure}
        \hfill
    \begin{subfigure}[b]{0.18\textwidth}
        \caption{69.88\%}
        \includegraphics[width=\textwidth,height=3.4cm]{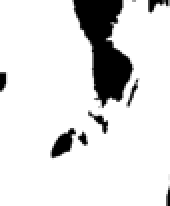}
        \caption{TransUNet}
    \end{subfigure}
        \hfill
    \begin{subfigure}[b]{0.18\textwidth}
        \caption{75.31\%}
        \includegraphics[width=\textwidth,height=3.4cm]{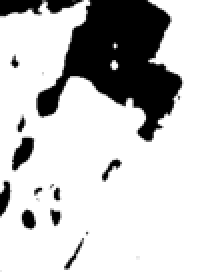}
        \caption{UNETR}
    \end{subfigure}
    \hfill
    \begin{subfigure}[b]{0.18\textwidth}
        \caption{77.77\%}
        \includegraphics[width=\textwidth,height=3.4cm]{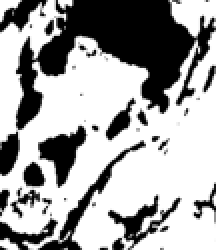}
        \caption{APF-UNETR}
    \end{subfigure}
    \caption{Example of segmentation quality for PAIP dataset. From $4K^2$ to $64K^2$ we zoom-in to show a portion of the image.}
    \label{fig:quality_res}
\end{figure*}

We demonstrate the quality of segmentation at different resolutions using different models: TransUNet, U-Net, UNETR, and our proposed APF-UNETR. We summarize the real results of the mask and display them in Figure~\ref{fig:quality_res}. The first column shows the original input, where the label is the resolution and the scaling percentage we use to show a portion of the image.

The second column shows the ground truth, followed by the prediction results of different models. It can be seen that at $512$ resolution, small patches cannot fully express the subtle differences. However, for high-resolution images, the deviations in subtle details will become larger and larger. 
At higher resolutions, uniform grid patching can only allow for very large patch sizes, such as $16K^2$ patch size with UNTER at input image of $64^2$ resolution. 
However, at the same input image resolution of $64K^2$, APF-UNETR, can still use patch sizes as small as $8^2$ in areas of detail by having more depth in the tree. This is the core benefit of adaptive patching.

\subsubsection{\textbf{Classification: APF vs. HIPT~\cite{Chen22}}} 
To demonstrate the versatility of APF, we compare classification for the PAIP dataset with the top performing and most sophisticated hierarchical multi-resolution model designed specially for microscopic pathology classification: HIPT~\cite{Chen22}. In this experiment, we divided the PAIP dataset, designed originally for segmentation, into six categories according to organs. Each category contains 40 samples, 28 of which are used for model training, 8 for testing, and 4 for validation. For HIPT, we resize all samples to three resolution scales $[256, 1024, 16384]$ and set the patch size for each scale to $[16, 256, 4096]$ according to the original settings. For the APF method, we only applied a level $16,384$ image for the classification; instead of using a decoder for segmentation work, we added an additional output channel for the class prediction. As seen in Table~\ref{table:classification_result}, with the same compute budget, using AFP with a vanilla ViT gains a huge improvement in accuracy ($>$7\%) over the very well-tuned and highly customized HIPT. At high-resolution ($16K^2$), the smallest patch size HIPT can handle, before going OOM, is $4,096^2$. AFP on the other hand can go down to patches of size $2^2$ at the regions of highest resolution in the images. This big gain in accuracy, despite using a vanilla ViT with AFP, indicates: a) the effectiveness of AFP, and b) that smaller patch sizes matter more than the sophistication of the model.

\subsection{Discussion}\label{sec:analysis}
\begin{table}[t]
	\centering
   	\caption{Classification (Top-1 accuracy) of vanilla ViT, HIPT~\cite{Chen22}, and APF-ViT on PAIP dataset ($16,384^2$ res.)}
	\begin{tabular}{l||c|c|c}
		\toprule[1.3pt]	
		\hline
		\thead{\textbf{Model}} & \thead{\textbf{Num. GPUs}} & \thead{\textbf{Patch Size}} & \thead{\textbf{Accuracy}} \\
            \hline
            ViT~\cite{dosovitskiy2021image} & 
            128 & 4,096 &  68.97  \\
            \hline
            HIPT~\cite{Chen22} & 
            128 & [16,256,4096] & 72.69 \\
            \hline
            APF-ViT-4096 & 
            8 & 4096 &  67.73  \\
            \hline
            APF-ViT-2 & 
            128 & 2 &  \textbf{79.73}  \\
		\hline
		\bottomrule[1.3pt]
	\end{tabular}
	\label{table:classification_result}
\end{table}

\subsubsection{\textbf{Adaptive Patches Empirical Growth Complexity}}
\begin{figure*}
    \begin{subfigure}[b]{0.32\textwidth}
        \includegraphics[width=\textwidth]{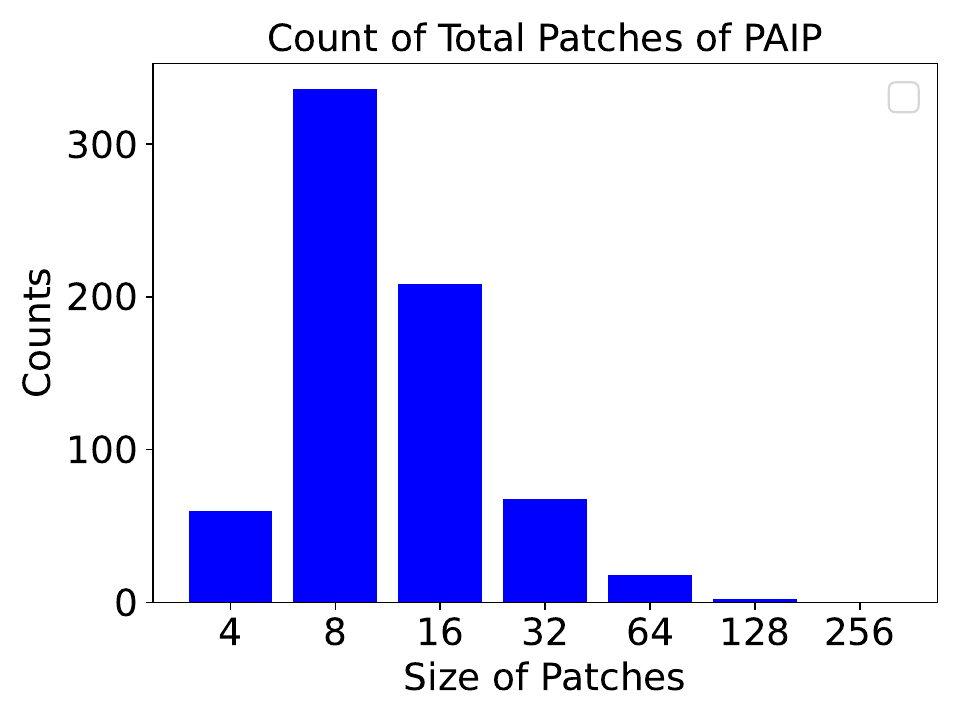}
        \caption{$v=20$, Avg patch size=9.37}
    \end{subfigure}
    \hfill
    \begin{subfigure}[b]{0.32\textwidth}
        \includegraphics[width=\textwidth]{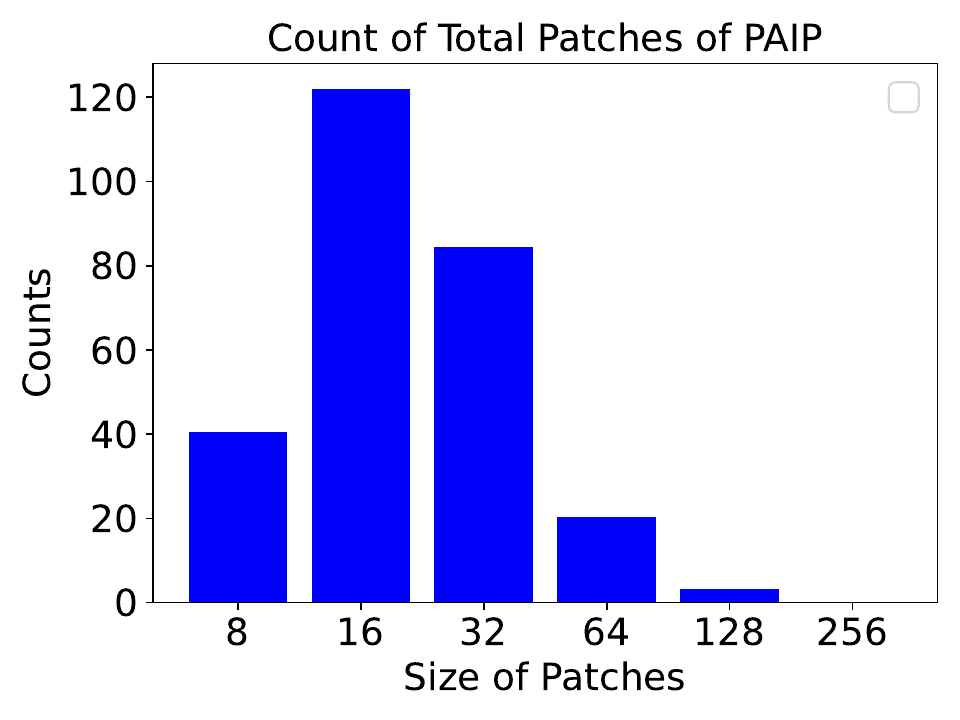}
        \caption{$v=50$, Avg patch size=20.21}
    \end{subfigure}
        \hfill
    \begin{subfigure}[b]{0.33\textwidth}
        \includegraphics[width=\textwidth]{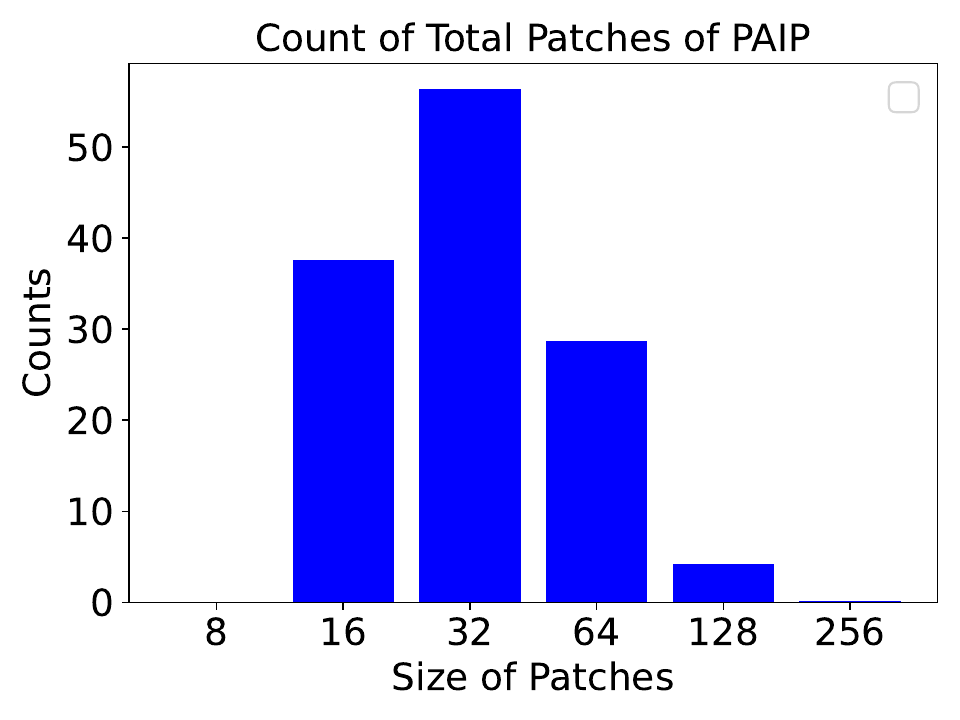}
        \caption{$v=100$, Avg patch size=30.73}
    \end{subfigure}
    \hfill
    \begin{subfigure}[b]{0.32\textwidth}
        \includegraphics[width=\textwidth]{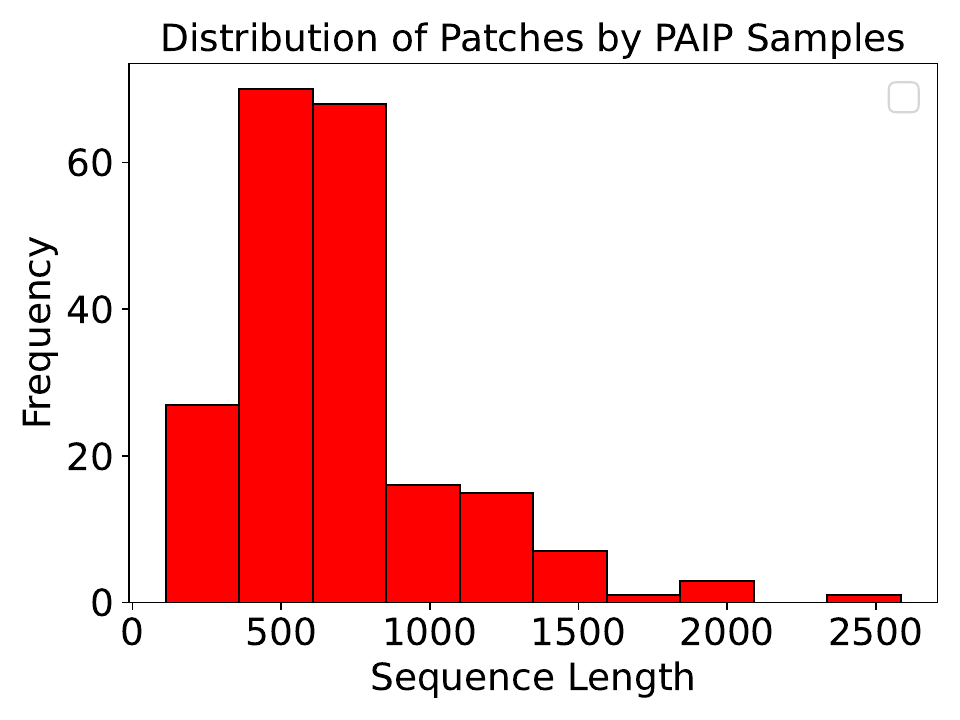}
        \caption{$v=20$, Avg length=677.7}
    \end{subfigure}
    \hfill
    \begin{subfigure}[b]{0.32\textwidth}
        \includegraphics[width=\textwidth]{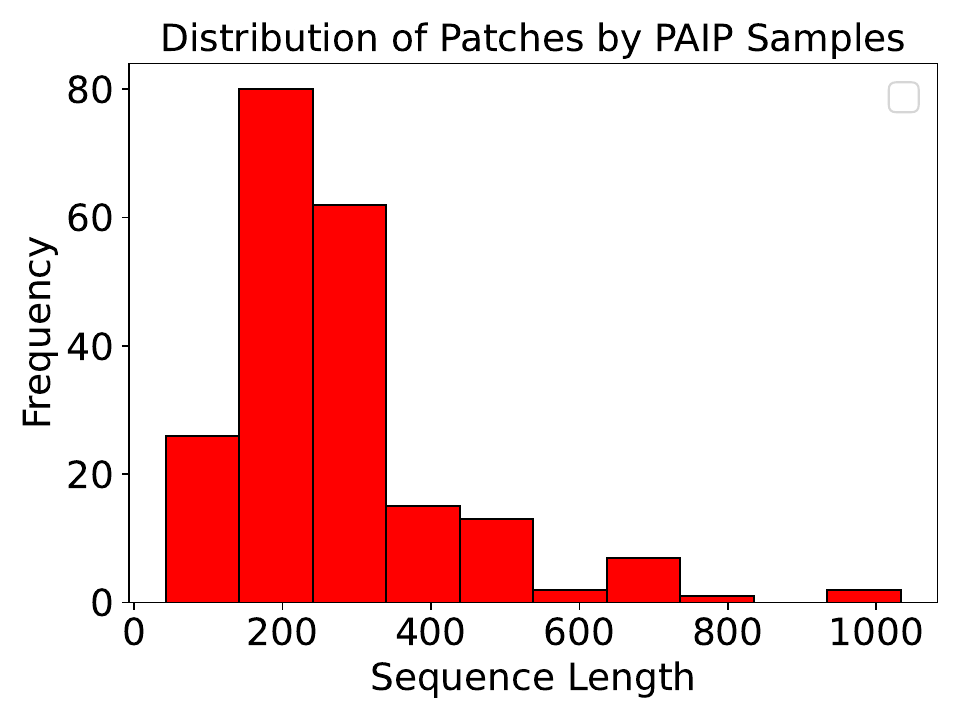}
        \caption{$v=50$, Avg length=286.9}
    \end{subfigure}
        \hfill
    \begin{subfigure}[b]{0.33\textwidth}
        \includegraphics[width=\textwidth]{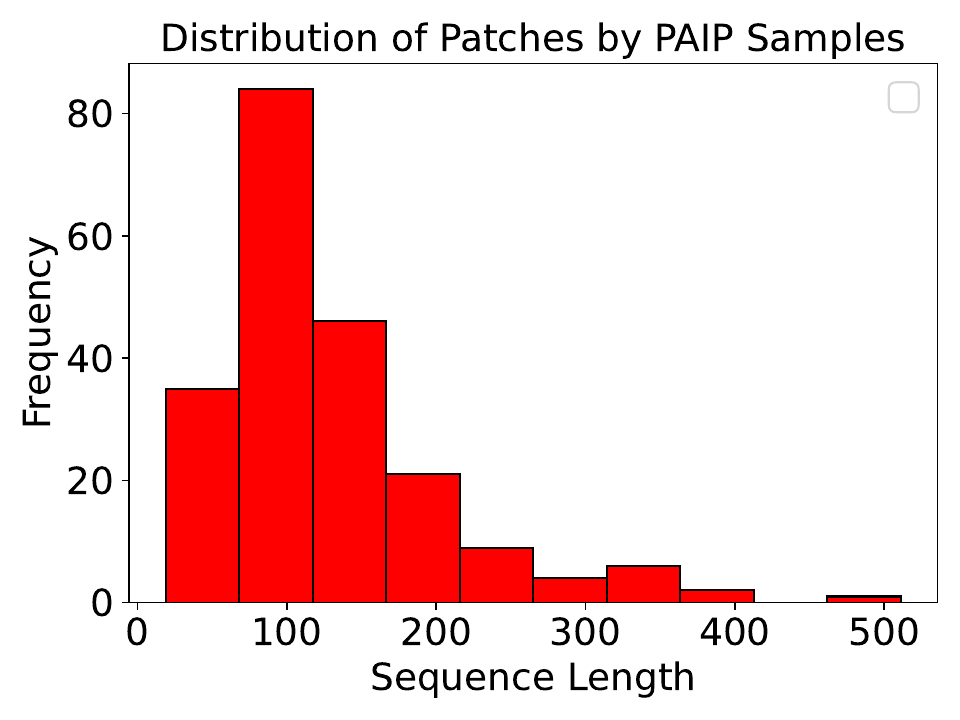}
        \caption{$v=100$, Avg length=127.5}
    \end{subfigure}
    \caption{Average quadtree patch size $[9.37, 20.21, 30.73]$ of training images in PAIP lead to empirical linear scaling of the corresponding average sequence length $[677.7, 286.9, 127.5]$, for different split values $[20, 50, 100]$. }
    \label{fig:p_v}
\end{figure*}
The core reason why APF can handle small patch sizes at high resolutions is that the sequence size can be reduced with adaptive meshing. In Figure~\ref{fig:p_v}, we show the extent to which the sequence length can be reduced by adjusting the split value of the quadtree, without significantly losing prediction performance. The split value $v$ controls the total length and distribution of patch sizes. The first row in Figure~\ref{fig:p_v} shows that when the split value is halved $[100, 50, 20]$, the patch size distribution or the average patch size $[9.37, 20.21, 30.73]$ is also close to being halved. This means the average patch size grows linearly with the split value. For the uniform grid patching strategy, the sequence length grows by $O(\frac{Z}{P})^2$. However, we observed an approximately linear increase in the average sequence length as the average patch size decreased. Note that APF sequence length depends on the complexity of the image itself, while the best case attention complexity is $O(log^2 N)$, the worst case would be $O(N^2)$ (it becomes like uniform grid patching). 

\subsubsection{\textbf{Training Stability and Patch Size}}
\begin{figure*}
    \begin{subfigure}[b]{0.32\textwidth}
        \caption{U-Net}
        \includegraphics[width=\textwidth]{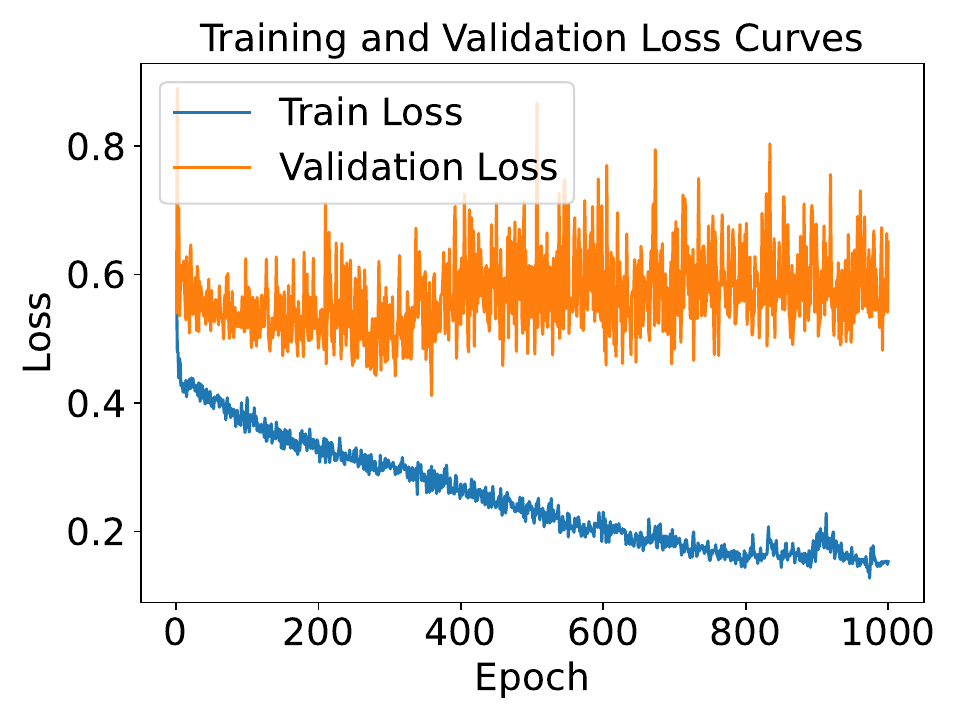}
    \end{subfigure}
    \hfill
    \begin{subfigure}[b]{0.32\textwidth}
        \caption{UNETR}
        \includegraphics[width=\textwidth]{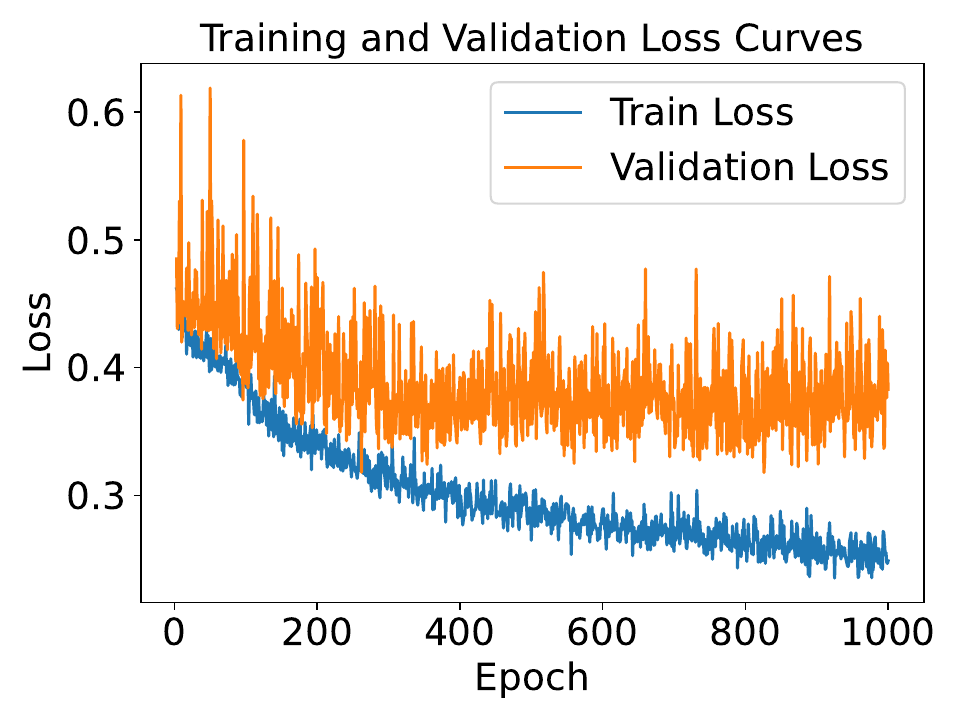}
    \end{subfigure}
        \hfill
    \begin{subfigure}[b]{0.32\textwidth}
        \caption{APF-UNETR}
        \includegraphics[width=\textwidth]{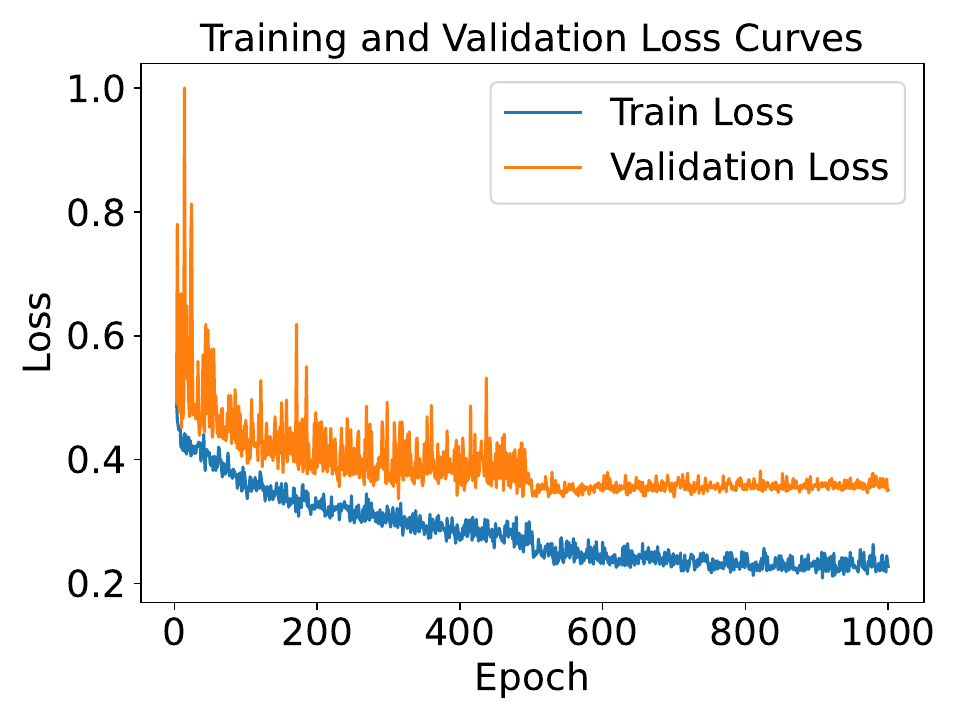}
    \end{subfigure}

    \begin{subfigure}[b]{0.32\textwidth}
        \includegraphics[width=\textwidth]{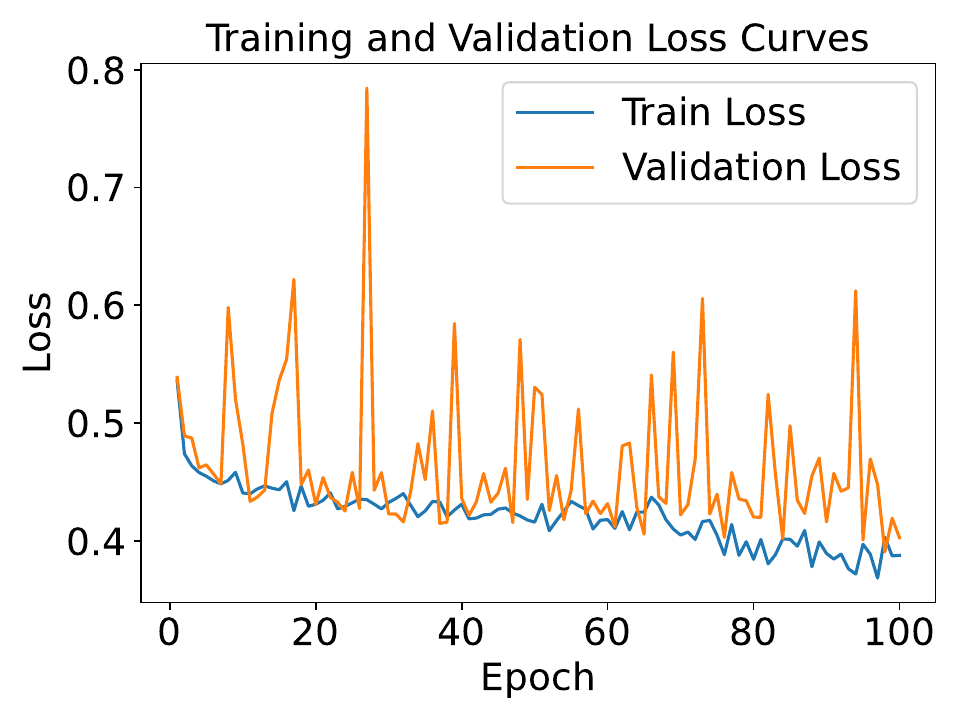}
        \caption{UNETR: $64 \times 64$ patches}
    \end{subfigure}
    \hfill
    \begin{subfigure}[b]{0.32\textwidth}
        \includegraphics[width=\textwidth]{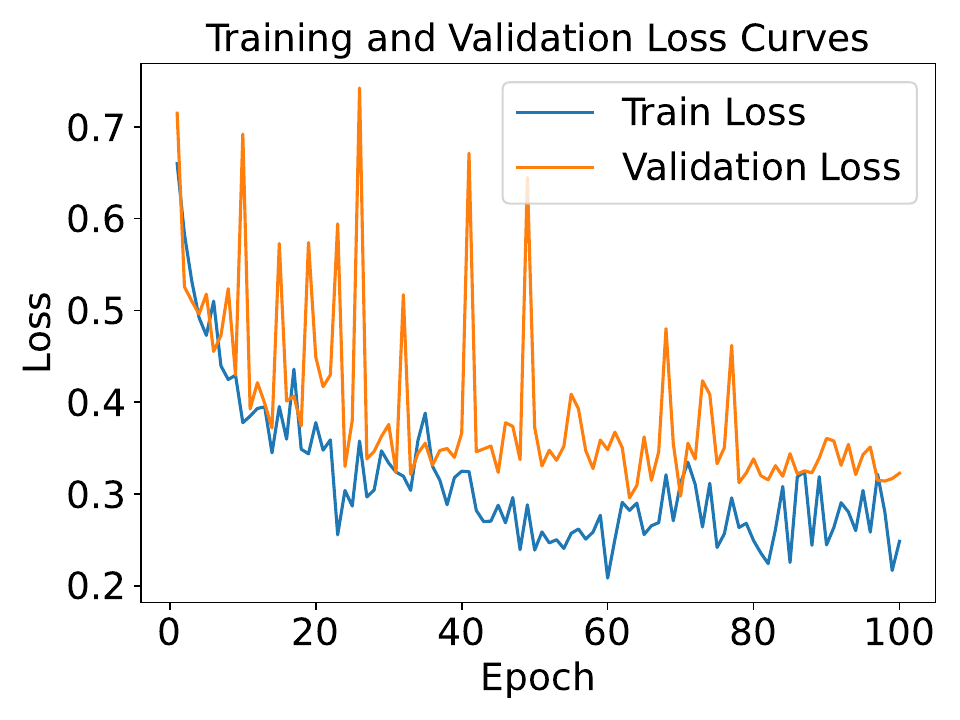}
        \caption{UNETR: $16 \times 16$ patches}
    \end{subfigure}
        \hfill
    \begin{subfigure}[b]{0.33\textwidth}
        \includegraphics[width=\textwidth]{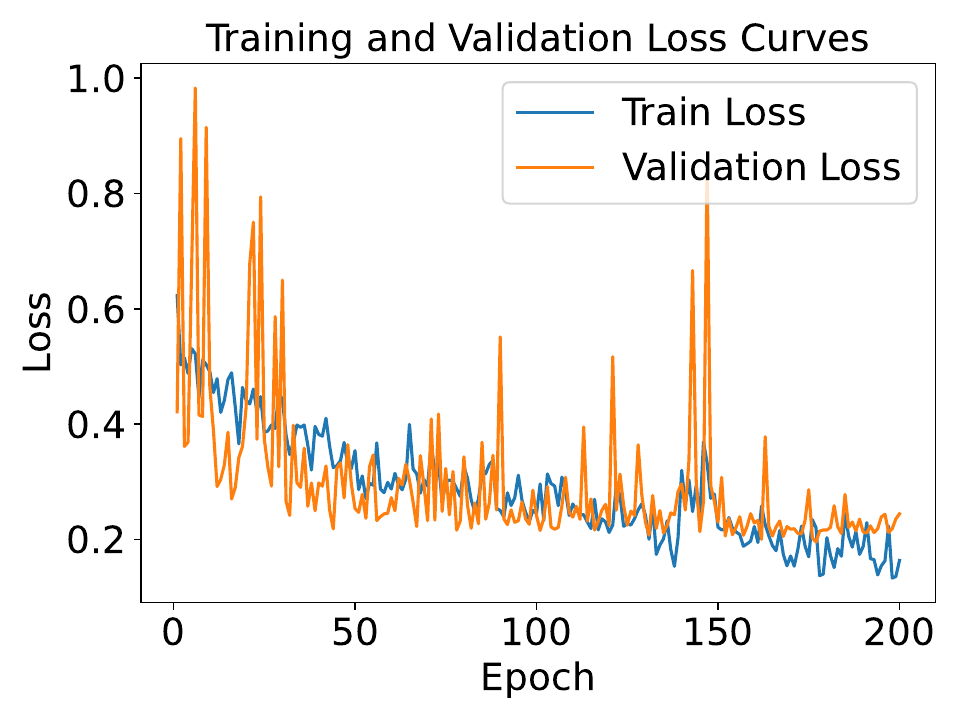}
        \caption{UNETR: $4 \times 4$ patches}
    \end{subfigure}
    \caption{Training and validation loss. (Top) different models. (Bottom) UNTER with different patch sizes.}
    \label{fig:curve}
\end{figure*}
Figure~\ref{fig:curve} shows the training and validation curves of the models: U-Net, UNETR-32 (patch size = 32), and APF-UNETR-2 (min. patch size = 2) at $4K^2$ resolution. We can see that at the same resolution and model complexity, the UNETR model using APF can converge to a better solution that is more stable than U-Net and UNETR. We hypothesize this is because APF allows the same model to use a smaller patch size under the same model complexity. To test this hypothesis, we further tried the performance of the UNETR model with different patch sizes $[4 \times 4 , 16 \times 16, 64 \times 64]$ at $1K^2$ resolution. The results further confirmed our thoughts. In Figure~\ref{fig:curve} (d,e,f), the UNETR model using a smaller patch size $4 \times 4$ tends to converge more stably than the bigger patch size $64 \times 64$.



\subsubsection{Overhead of APF: Negligable}
In our experiments, the time is taken for the PAIP dataset with resolutions $[512,1024,4096,32768,65536]$ is $[4.232, 7.561, 37.160, 127.374, 286.568]$ in seconds. This is negligible when compared with training time (Hours).

\section{Conclusion}
We propose a solution that adaptively patches high-resolution images based on image details, drastically reducing the number of patches fed to vision transformer models. This pre-processing approach incurs minimal overhead. We achieve segmentation quality for $64K^2$ images comparable to SoTA models operating on no more than $4K^2$, at much higher efficiency (geomean speedup of $6.9\times$).



\bibliographystyle{IEEEtran}
\bibliography{conference_101719}

\end{document}